\documentclass[10pt,journal]{IEEEtran}
\usepackage[utf8]{inputenc}
\usepackage{outlines}
\usepackage{xcolor}
\usepackage{graphicx}
\usepackage[normalem]{ulem}
\usepackage{array}
\usepackage{tikz}
\usetikzlibrary{positioning}
\usetikzlibrary{calc}
\usepackage{forest}
\usepackage{pgfplots}
\usepackage{pgfplotstable}
\pgfplotsset{compat=1.17}
\usepackage{booktabs}
\usepackage{url}
\usepackage{amsmath}
\usepackage{amssymb}
\usepackage{placeins}
\usepackage{outlines}
\usepackage{floatrow}
\usepackage{subfig}
\usepackage{enumitem}

\usepackage{pifont}

\usepackage[colorlinks=true,linkcolor=black,anchorcolor=black,citecolor=black,filecolor=black,menucolor=black,runcolor=black,urlcolor=black]{hyperref}
\usepackage[nocompress]{cite}
\usepackage{orcidlink}

\title{DSAP: Analyzing Bias Through Demographic Comparison of Datasets}
\author{Iris~Dominguez-Catena\orcidlink{0000-0002-6099-8701},~\IEEEmembership{Student Member,~IEEE,} 
Daniel~Paternain\orcidlink{0000-0002-5845-887X},~\IEEEmembership{Member,~IEEE,}
Mikel~Galar\orcidlink{0000-0003-2865-6549},~\IEEEmembership{Member,~IEEE}%
\IEEEcompsocitemizethanks{\IEEEcompsocthanksitem I.Dominguez-Catena, D. Paternain and M. Galar are with the Department of Statistics, Computer Science and Mathematics, Public University of Navarre (UPNA), Arrosadia Campus, 31006, Pamplona, Spain and with the Institute of Smart Cities (ISC), Public University of Navarre (UPNA), Arrosadia Campus, 31006, Pamplona, Spain.\protect\\
E-mail: iris.dominguez@unavarra.es; daniel.paternain@unavarra.es; mikel.galar@unavarra.es}
\date{2023}

\thanks{This work was funded by a predoctoral fellowship from the Research Service of the Universidad Publica de Navarra, the Spanish MICIN (PID2019-108392GB-I00,  PID2020-118014RB-I00 and PID2022-136627NB-I00 / AEI / 10.13039/501100011033), and the Government of Navarre (0011-1411-2020-000079 - Emotional Films).
}}

\begin{document}

\maketitle
\begin{abstract}

    In the last few years, Artificial Intelligence systems have become increasingly widespread. Unfortunately, these systems can share many biases with human decision-making, including demographic biases. Often, these biases can be traced back to the data used for training, where large uncurated datasets have become the norm. Despite our knowledge of these biases, we still lack general tools to detect and quantify them, as well as to compare the biases in different datasets. Thus, in this work, we propose DSAP (Demographic Similarity from Auxiliary Profiles), a two-step methodology for comparing the demographic composition of two datasets. DSAP can be deployed in three key applications: to detect and characterize demographic blind spots and bias issues across datasets, to measure dataset demographic bias in single datasets, and to measure dataset demographic shift in deployment scenarios. An essential feature of DSAP is its ability to robustly analyze datasets without explicit demographic labels, offering simplicity and interpretability for a wide range of situations. To show the usefulness of the proposed methodology, we consider the Facial Expression Recognition task, where demographic bias has previously been found. The three applications are studied over a set of twenty datasets with varying properties. The code is available at \url{https://github.com/irisdominguez/DSAP}.

\end{abstract}


\section{Introduction}

    The development of Artificial Intelligence systems in recent years has been characterized mainly by the creation of large models based on Deep Learning techniques, such as transformers~\cite{Raffel2020} and diffusion models~\cite{Ho2020}. In these systems, the models are trained on large amounts of data, recognizing inherent patterns in those data. This ability to abstract general patterns from specific data allows these systems to generalize the knowledge obtained to new contexts and previously unobserved data.
    
    However, this learning approach based on increasingly large datasets also leads to datasets that are harder to curate, often filled with biased, harmful, and even outright false information~\cite{Beyer2020,Prabhu2020}. Despite the continuous efforts of the teams collecting these datasets~\cite{Schumann2021,Surabhi2022}, the size of the datasets makes it virtually impossible to guarantee that there is no offensive content, hate speech, misrepresentations, stereotypes, or other potentially harmful data patterns within them. All this information, especially when systemic and repeated throughout the dataset, leads to unwanted patterns that are difficult to distinguish from the acceptable ones for trained models~\cite{Dominguez-Catena2022}.
    
    These types of unwanted patterns are commonly known as biases and have been found not only in the source datasets, but also in several locations in the ML pipeline~\cite{Ntoutsi2020,Suresh2021}. All of these sources of bias can, in some way or another, be transferred to the final predictions of the model, which can lead to a differentiated treatment of users according to protected demographic attributes such as race, age, or gender~\cite{Dominguez-Catena2022}. Most current and planned legislation~\cite{Park2023} focuses mainly on these behaviors, regardless of the origin of the bias or discriminatory behavior of the model. Despite this myopic legal approach, research on the sources of biases, and particularly the bias that originated from unwanted patterns and underrepresentation in datasets, becomes crucial, as it enables the removal of bias early in the machine learning pipeline before propagation and supports mitigation approaches~\cite{Hort2022}.
    
    In the literature, dataset demographic biases have previously been studied using specific measures~\cite{Dominguez-Catena2023a}. Measured biases can be broadly classified into representational and stereotypical biases. While representational bias focuses on the degree of representation of the demographic groups studied relative to each other, stereotypical bias refers to the under or overrepresentation of specific demographic groups for a given target class. Different measures can be used to measure each type of bias, either in general or focusing on specific aspects or subtypes of these biases, such as the Effective Number of Species~\cite{Jost2006} for general representational bias, the Shannon Evenness Index~\cite{Pielou1966} for evenness, one of the key components of representational bias, and the Cramer's V~\cite{Cramer1991} for stereotypical bias. Unfortunately, since each measure is applicable only to one type of demographic bias with its own range and interpretability, it becomes difficult to compare the magnitude of the different types of biases in a single dataset, and even more between datasets. Despite this, the demographic comparison of datasets can be highly valuable in many applications. For example, when combining datasets~\cite{Gao2020}, it can help guide the selection of those that maximize demographic diversity; when studying biases in the datasets, it can serve to compare them with ideal versions; when monitoring deployed models, it can allow detecting demographic changes in the population using the system. However, to the best of our knowledge, no demographic dataset comparison methodologies have been proposed in the literature.

    \begin{figure*}[ht]
        \centering
        \includegraphics[width=\textwidth]{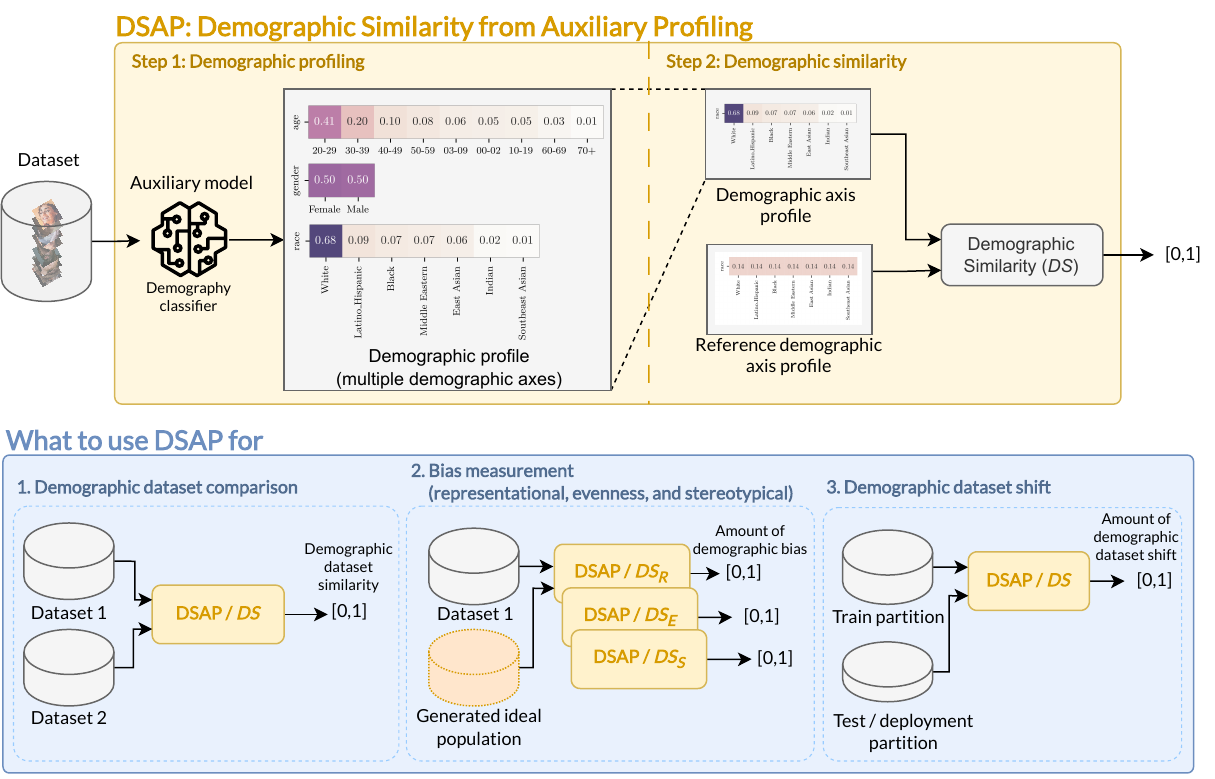}
        \caption{DSAP summary. The two steps illustrated at the top of the figure can then be applied for the three applications in the lower part.}
        \label{figure:dsap_summary}
    \end{figure*}
    
    Therefore, in this work, we propose DSAP (Demographic Similarity from Auxiliary Profiles), a two-step methodology to compare the demographic profiles of two datasets, and test its usefulness in three applications related to dataset bias. A summary of the methodology is shown in Figure~\ref{figure:dsap_summary}, along with its three general applications. The first step of DSAP consists of generating an approximation of the demographic characteristics of the subjects represented in the dataset, which we call the demographic profile. These demographic profiles consist of several demographic axis profiles, each one giving a view of the dataset according to a demographic characteristic, such as age, gender or race. Since most datasets do not contain demographic information or the information is not codified in compatible ways between datasets, in this first step we use an external auxiliary model that is capable of approximating demographic attributes for each sample and then aggregate that information to generate the profile. The second step focuses on measuring the similarity between two of these profiles for a given demographic axis. For this similarity measurement, we take inspiration from the fields of ecology and archaeology, where similar problems have already been studied and measures proposed~\cite{Wilson1984,Robinson1951}. Consequently, we propose the usage of the Renkonen similarity index~\cite{Renkonen1938}, which for this application is mathematically equivalent to other well-known measures such as the Sørensen-Dice index~\cite{Dice1945} or the Brainerd-Robinson Similarity Coefficient~\cite{Robinson1951}.
    
    Depending on the demographic axis profile used as the target for computing the similarity, DSAP has different applications: pairwise dataset comparison, dataset bias measurement, and data shift and drift measurement. The first application, direct pairwise comparison between datasets, can be used to investigate existing biases in datasets that are being used in a specific field, or to evaluate datasets that are intended to be combined~\cite{Gao2020} or used sequentially in pretraining and fine-tuning methodologies~\cite{Raffel2020}. For the second application, bias measurement, instead of comparing the input dataset with other datasets, we compare it with an ideal unbiased dataset. Furthermore, the ideal axis profiles of this unbiased dataset can follow specific target populations (different from balanced ones), which was not possible with previous demographic bias measures~\cite{Dominguez-Catena2023a}. This allows us to model real-world demographics, adapting our bias definition to more realistic targets (e.g., the uneven age distribution over the world). Finally, in its third application, DSAP can be used to measure demographic dataset shift, detecting changes in the demographic profile between training and test or deployment datasets, which can severely deteriorate model performance~\cite{Lu2018}.
    
    To test DSAP and show its usefulness, we focus on Facial Expression Recognition (FER)~\cite{Barsoum2016}. FER is a challenging task in which facial expressions (typically the 6 basic emotions defined by Ekman~\cite{Ekman1971}) are identified from a picture of a face. FER applications are varied, including healthcare, assistive robotics, marketing, and entertainment. In many of these applications, bias issues can easily go unnoticed, as end users are often unsupervised, and many systems do not provide explicit feedback. This task is affected by demographic biases in source datasets as shown in~\cite{Dominguez-Catena2022} and~\cite{Dominguez-Catena2023}. Therefore, it is a suitable use case to test our methodology. In this work, we perform a detailed analysis of our methodology in the three applications we devised using a selection of the twenty most relevant open datasets for FER.
    
    The following sections are as follows. First, Section~\ref{s:related} summarizes related works. Next, Section~\ref{s:methodology} presents the DSAP methodology and describes how to apply it to the different applications. Section~\ref{s:casestudy} describes the dataset selection process and details how we preprocess them. Section~\ref{s:results} details the findings of our experiments. Finally, Section~\ref{s:conclusion} concludes this work.

\section{Related work}\label{s:related}

    In this Section we summarize some previous works and related literature. First, Section~\ref{s:rel_bias} explains the concept of dataset bias and its relation to the more general concept of fairness. Next, Section~\ref{s:rel_comp} focuses on the topic of demographic comparison of datasets. Section~\ref{s:rel_bias_measures} reviews specific measures for dataset bias. Section~\ref{s:rel_shift} explains the idea of demographic dataset shift. Finally, Section~\ref{s:rel_fer} presents the task for our case study, FER.

\subsection{Fairness and dataset bias}\label{s:rel_bias}

    The widespread deployment of AI systems in recent years in various fields has led to increasing concern about their fairness and bias characteristics~\cite{Pessach2020,Landers2022}. Although fairness is a disputed and complex concept, without a unanimous definition, it is more commonly defined through specific examples of unfairness, called demographic biases. These biases are unwanted differences in the treatment or representation of people based on their demographic attributes, such as the degraded performance of FER systems in specific age groups~\cite{Kim2021}. Furthermore, this conception of fairness is directly related to the legal notion of bias~\cite{Ntoutsi2020}, where it is defined as a prejudice against a person or group of people based on some protected demographic attributes, usually inherent to the person and not chosen, such as race, gender, sex, age, religion, or socioeconomic class. With new regulations and ethical concerns pushing for the evaluation of fairness in most AI systems, the ability to analyze specific biases across these demographic axes is vital.

    Although most demographic bias analysis focuses on the final bias exhibited by the system, called model bias, it can be attributed to several independent sources found at several points in the ML pipeline~\cite{Suresh2021}. The tendency to have increasingly large data sets in ML, especially in image~\cite{Deng2009} and text~\cite{Gao2020} applications, compounded with the adoption of pretrained models trained on them~\cite{Howard2018,Long2022}, makes dataset bias one of the most relevant sources of bias.

    Different types of dataset bias can arise depending on the type of data contained in a dataset. For example, in image datasets bias can be categorized in selection, framing, or label bias~\cite{Fabbrizzi2022}. From these categories, selection bias refers to the way populations are statistically under or overrepresented in the dataset, while framing and label bias refer to the contextual information in the image and to the target variables, respectively. Although label and framing bias can strongly impact the fairness of models, they tend to be highly dependent on the task, while selection bias issues can arise on any dataset that includes human subjects, even in non-image based datasets. Therefore, in this work, we focus primarily on dataset bias from the perspective of selection bias, working only with the number of samples corresponding to each demographic group. This type of dataset bias can be further divided into two subtypes:

    \begin{itemize}
        \item \textbf{Representational bias}. This type of bias is a direct over or underrepresentation of demographic groups in a dataset. It can be divided into several components or subtypes, namely, \textit{richness}, \textit{evenness}, and \textit{dominance}. From these, richness focuses on the raw number of demographic groups represented, evenness on how equally or unequally represented they are, and dominance on how much the most represented group dominates over the rest of the dataset. For example, in FER datasets, it is common to have an overrepresentation of white people with respect to other races.
        \item \textbf{Stereotypical bias} identifies when demographic groups have correlations with the target variable,  thus being a mixture of selection and label bias. In the case of FER, for example, a common stereotypical bias is an overrepresentation of female-presenting people in the happy class, while they tend to be underrepresented in the angry class.
    \end{itemize}
    
\subsection{Demographic dataset comparison}\label{s:rel_comp}

    The idea of comparing datasets is common in the literature, where most new datasets are in some way compared to previous ones in characteristics such as size, composition, or data source. Despite this, comparison at the demographic level is much less common, reserved for datasets that explicitly deal with demography~\cite{Banziger2012,Guerdelli2022,Karkkainen2021}, with very few studies focused on demographic comparison itself~\cite{RobnikSikonja2018,Coutinho-Almeida2022}. When presenting new datasets, demography is generally summarized in statistics and not directly compared~\cite{Banziger2012}, or compared at the global statistic level~\cite{Guerdelli2022}. When a more exhaustive comparison is made~\cite{Karkkainen2021}, it is usually limited to a single demographic attribute, manually comparing the relative proportions of each possible group between datasets. Most of the work explicitly dedicated to dataset comparison deals only with tabular data~\cite{RobnikSikonja2018,Coutinho-Almeida2022} which, although easier to compare and investigate, have little resemblance to the image and text datasets currently used for many tasks. In tabular datasets, protected attributes and demographic information are usually available, whereas for many image- and text-based tasks, that information is only available through proxies. Despite the few works that deal with demographic dataset comparison in Machine Learning, looking for new ways of comparing populations is important to gain a deeper understanding of the existing biases and help mitigate them.

    To do so, we can draw inspiration from similar tasks that have already been studied in other research fields such as archaeology~\cite{Robinson1951} or ecology~\cite{Wilson1984,Ricotta2017}. In these fields, specific similarity measures and techniques have already been developed that can be adapted to demographic dataset comparison.
    
    In archaeology, artifacts found at different sites are classified, generating an artifact profile, i.e., a distribution of artifact classes, which can be interpreted as a demographic profile. The profiles are then compared using a measure such as the Brainerd-Robinson similarity coefficient~\cite{Robinson1951}, with the objective of chronologically ordering the sites. Furthermore, this measure deals with the proportion of each type of artifact in the site, instead of using the total counts, which is well suited to the problem of comparing demographic profiles between datasets, which requires accounting for the high variability in dataset sizes.
   
    In ecology, similar measures are used to measure the diversity between ecosystem populations on different scales. In particular, the ecological concept of $\beta$-diversity refers to the differentiation between samples collected at different sites. A popular $\beta$-diversity measure is the Renkonen similarity index~\cite{Renkonen1938}, which when applied to relative populations is mathematically equivalent to several other well-known measures, such as the Bray-Curtis similarity/dissimilarity~\cite{Ricotta2017}. Under these constraints, the measure is also mathematically equivalent to the Brainerd-Robinson similarity coefficient scaled to the unitary range. Although there are more popular measures for $\beta$-diversity, most of them deal with the presence / absence data~\cite{Wilson1984} or with the absolute number of subjects found in each species. Unfortunately, in the context of demographic dataset comparison, the sizes of the datasets vary greatly. This may result in the presence of a heavily underrepresented group in a large dataset, which is absent in a smaller dataset just for statistical reasons. This makes presence/absence data unreliable and makes using relative-population-based profiles a more suitable approach. Current best practices in ecology have even evolved beyond these group representation measures, incorporating information about the relatedness of the different groups~\cite{Lozupone2005}. In the context of dataset demographic comparison, unfortunately, establishing relatedness scores between demographic groups is nearly impossible.
    
\subsection{Bias measures}\label{s:rel_bias_measures}

    To unify the notation between the measures presented in this section and our proposals in Section~\ref{s:met_measure}, we employ the following conventions.

    \begin{itemize}
        \item We define $G$ as the set of demographic groups defined by a specific protected attribute, which can take a finite and predefined number of values. For example, if $G$ stands for \textit{gender presentation}, a possible set of groups would be $G = \{\text{masculine}, \text{feminine}, \text{androgynous}\}$.
        \item We define $Y$ as the set of target classes of a problem.
        \item We define $X$ as a population of $n$ samples.
        \item We define $n_g$ as the number of samples from the population $X$ that belong to the demographic group $g \in G$.
        \item Similarly, we define $p_g$ as the proportion of samples from the population $X$ that belong to $g$: $$p_g = \frac{n_g}{n}\;.$$
    \end{itemize}

    Previous works~\cite{Dulhanty2019,Buolamwini2018,Zhao2017,Kim2019,Dominguez-Catena2023a} have explored metrics for most of the types of bias presented in Section~\ref{s:rel_bias}. A previous work~\cite{Dominguez-Catena2023a} provides an overview of the different measures of demographic bias and empirically determines a sufficient set of measures to describe demographic selection bias. In this work, we take these measures as a reference, which are recalled hereafter.

    \textbf{Effective Number of Species} (ENS)~\cite{Jost2006}. ENS is a measure of representational bias, combining both of its key components: richness (the raw number of represented groups) and evenness (the homogeneity of their representation). This measure has an unusual range between $1$ (for datasets with only one group) and the number of groups represented (for completely balanced datasets). This improves its interpretability at the cost of losing any comparability with other measures, which are defined on different domains. It is defined as:

    \begin{equation}
        \text{ENS}(X) = \exp\left({-\sum_{g\in G}{p_g \ln p_g}}\right)\;.
    \end{equation}

    \textbf{Shannon Evenness Index} (SEI)~\cite{Pielou1966}. SEI is a measure focused on evenness, that is, the homogeneity of the represented groups, independently of the number of represented groups. This measure is bounded in the unitary range, with a value close to $0$ for the more uneven (biased) datasets, and close to $1$ for the more balanced ones. It is defined as follows:
    
    \begin{equation}
        \text{SEI}(X) = \frac{\ln{(\text{ENS}(X))}} {\ln(\text{Rich}(X))}\;,
    \end{equation}
    where $\text{Rich}(X)=|\{n_g \text{ such that }n_g>0\}|$ is the number of represented groups in $X$.



    \textbf{Cramer's V} (V)~\cite{Cramer1991}. Finally, Cramer's V is a measure of stereotypical bias, based on the correlation between the target variable and a demographic variable. It is bounded in the unitary range, in this case with $0$ corresponding to unbiased datasets, and $1$ to stereotypically biased ones. It is defined as follows:

    \begin{equation}
        \text{V}(X) = \sqrt{ \frac{\chi^2(X)/n}{\min(|G|-1,|Y|-1)}}\;,
    \end{equation}
    where $\chi^2(X)$ is the Pearson's ${\chi^2}(X)$ statistic, defined as:
    \begin{equation}
        \chi^2(X)=\sum_{g \in G}\sum_{y \in Y}\frac{(n_{g\land y}-\frac{n_g n_y}{n})^2}{\frac{n_g n_y}{n}}\;.
    \end{equation}

\subsection{Demographic dataset shift}\label{s:rel_shift}

    The concept of dataset shift refers to a variation in the distribution of the characteristics of the data used in training compared to the data used to evaluate or deploy the model~\cite{Bayram2022}, although the concept has been studied under multiple denominations and in slightly different contexts~\cite{Moreno-Torres2012}. Dataset shift can impact either the evaluation of the model, or its deployment performance~\cite{Gama2004}. 
    
    When demographic variations occur between the training and test partitions of the model, test metrics can over- or underestimate the performance of the final model. Often, this problem could be avoided computing independent metrics for each demographic group, but this is not common practice in the literature. Furthermore, it would become impractical for large numbers of demographic groups. Otherwise, when these demographic variations occur between the training dataset and the data encountered in a production scenario (i.e., model drift), the model can become unreliable making inaccurate predictions, for example, for previously unseen demographic groups.
    
    Previous work on data shift~\cite{Moreno-Torres2012} has mainly focused on measuring different types of shifts attending to the variation of one or more input variables, for example in data streaming scenarios~\cite{Webb2016}. In our case, we focus only on the variations of protected attributes or their proxies. To the best of our knowledge, no other work has measured the shift in demographic characteristics or their proxies, an indispensable tool to detect demographic biases that appear after deployment.

\subsection{Facial Expression Recognition}\label{s:rel_fer}

    FER~\cite{Assuncao2022} is one of the most common tasks in the interdisciplinary field of affective computing. Although there are several modalities for general emotion recognition, recognizing facial expressions from face photos is a typical approach, as it is relatively simple to implement, requiring minimal hardware and easy to obtain data. Despite this, it is also a challenging problem that has only become possible thanks to the developments of Deep Learning techniques~\cite{Barsoum2016}, characterized by the use of large datasets to train large models.

    For FER tasks, either the expressions themselves or the underlying emotional states have to be codified and annotated. Although detailed expressions and facial features can be identified, the use of a restricted set of stereotypical expressions makes data labeling easier. For this reason, many large datasets, especially those crowd-labeled from internet searches, use a scheme based on the six basic emotions defined by Ekman~\cite{Ekman1971}, each associated with their stereotypical expressions.

    Unfortunately, the large datasets used for FER are not free from the biases described in Section~\ref{s:rel_bias}. In particular, it has been shown that most datasets are affected by both representational and stereotypical demographic biases~\cite{Dominguez-Catena2023a}, and that these biases transfer and modify the behavior of the final model~\cite{Dominguez-Catena2022,Dominguez-Catena2023a}. These biases have been found in the demographic axes of age, gender, and race, since they have strong influences on the facial images gathered in these datasets. Nonetheless, other demographic attributes can also alter facial expression performance and recognition, such as social class~\cite{Bjornsdottir2017} or drug use~\cite{Raghavendra2016}.

\section{DSAP: Demographic Similarity from Auxiliary Profiling}\label{s:methodology}

    This section presents our proposed DSAP methodology for comparing the demographic profiles of two datasets, together with the details of its application to the three proposed use cases. DSAP methodology is divided into two sequential steps (see Fig. \ref{figure:dsap_summary}):
    \begin{enumerate}
        \item \textbf{Demographic profiling: } The demographic profile of a given dataset is obtained through an auxiliary model capable of predicting demographic properties with a certain degree of accuracy.
        \item \textbf{Demographic similarity: } A similarity measure is applied between one of the demographic axes of the extracted demographic profile (e.g., age or gender) and a reference demographic profile. The reference profile could either be a profile extracted from another dataset or a profile generated manually.
    \end{enumerate}
    As a result of this methodology, we are able to provide a single intuitive value that summarizes the demographic similarity between the input dataset and the reference profile on a given demographic axis. These two steps are described in  detail in Section~\ref{s:met_profiling} and Section~\ref{s:met_measure}, respectively. 
    
    Furthermore, based on this methodology, we derive three applications of DSAP, which mainly vary depending on what is used as a reference profile (see also Fig. \ref{figure:dsap_summary}). These applications are described in Section \ref{s:met_comparison} (demographic dataset comparison), Section \ref{s:met_bias} (bias measurement), and Section \ref{s:met_shift} (demographic dataset shift measurement).
    

    

\subsection{Step I: Demographic profiling}\label{s:met_profiling}

    Most large datasets, especially those created using data from the Internet, that is, freely available and uncontrolled data, do not include additional information about the samples besides the target variable. Except for a few data sets that are explicitly designed for the classification of demographic attributes~\cite{Karkkainen2021,Cao2018}, where demographic labels are available directly, most datasets do not have explicit demographic information. However, these attributes can leak into observable proxies~\cite{Chen2019}, and auxiliary models can estimate the original demographic characteristics of the subjects. Even when demographic information is available, it is not always codified in compatible taxonomies or groups between different datasets, or they may follow different annotation processes, making comparisons difficult or directly impossible.
    
    For these reasons, we propose a first step of demographic profiling based on a common auxiliary model. This model or models should be able to obtain a demographic profile with the relevant information from the observable proxy variables (in the case of image classification, for example, the whole image acts as a single proxy variable). Employing the same model for all datasets, we guarantee that the profiles are compatible and that any potential bias in the model affects all datasets in a similar way.

    The choice of the auxiliary model and the target protected attributes to be studied is mainly determined by the input information in the dataset, and the reliability of the potential observable proxies. For example, in the case of FER, the input facial images can easily encode gender, sex, age, and ethnicity. Gender and sex are usually entangled in gender expression choices such as clothing or makeup usage, and also reflect in some secondary sexual characteristics. Age can be codified more directly in facial features, although with a fair amount of variability. Finally, some ethnic background can be identified in clothing clues, hairstyles, and ethnic phenotypical characteristics, such as skin color, and can usually be summarized into a broad race category. Although all of these proxy categories are naturally imprecise, if they are significantly biased in the dataset, they are susceptible to discrimination issues in the final model~\cite{Dominguez-Catena2022}.

    Regarding the taxonomy and codification of demographic attributes into variables, in this work we focus on nominal variables, that is, based on finite and unordered demographic groups. Some models and demographic axes can support other types of variable, such as ordinal and continuous codifications of a person's age. Although the additional information codified in them can provide a more exact comparison, the approximate nature of a demographic profile based on an auxiliary model usually does not support that level of precision. Moreover, using the same type of variable for all possible cases makes the cross-axes comparison easier to interpret. Additionally, we can generate new axes from the combination of the groups in the others, obtaining new perspectives fro demographic subgroups. For these combination axes, we simply take two or more independent demographic axes and consider each possible combination of demographic groups in them as a single group. It is important to note that for combination axes, the possible number of groups increases substantially, which can alter the distribution of any derived measurements.

    Although different potential auxiliary models are available, in this work, we use FairFace~\cite{Karkkainen2021}, a demographic classification model that predicts age, gender, and race. As our profiles are derived from this model, it also determines the particular codification of the demographic properties that we use and the demographic axes that we consider, that is, age, gender, race, and the additional combined axis of these three. See Section~\ref{s:cs_fairface} for additional details.

    As a result of applying the auxiliary model to a dataset, we obtain the full list of demographic properties for each sample, which is then aggregated to obtain the demographic profile of the dataset in each studied axis. However, although the total counts in the dataset play an important role in training a model, our notion of dataset bias is more related to the disproportion between the representation of the demographic groups, which makes working with relative sizes more suitable for bias measurement and related tasks. Therefore, we consider the demographic profile for each axis as the relative population of each group on that demographic axis in the data set.\footnote{We refer to demographic profile as the overall demographic properties of the samples in a dataset, whereas we refer to demographic axis profile as the specific relative counts of the different demographic groups in a given axis (e.g., age or gender), which are the inputs for the second step of DSAP.}

\subsection{Step II: Demographic similarity}\label{s:met_measure}

    Once we have extracted the demographic profile from a given dataset, the objective of this second step is to obtain a single intuitive value representing the similarity between this profile and a reference profile, measured on a specific demographic axis. Recall from Section~\ref{s:rel_comp} that different measures are available to compare populations. When comparing normalized demographic profiles, as in this case, most of the applicable measures can be classified into two families, Renkonen and Jaccard.
    
    Measures from the Renkonen family include the Renkonen similarity index itself~\cite{Renkonen1938}, the quantitative Sørensen-Dice~\cite{Dice1945}, Whittaker~\cite{Whittaker1952}, Czekanowsky~\cite{Czekanowski1909} and Bray Curtis~\cite{Ricotta2017} similarity indexes. Additionally, the Brainerd-Robinson similarity index~\cite{Robinson1951} also falls into this category when normalized to the unit range. Following the notation presented in Section~\ref{s:rel_bias_measures}, the unified definition for these metrics is:

    \begin{equation}
        \text{R}(X, X') = \sum_{g\in G} \min \left(p_g, p'_g\right)\;.
    \end{equation}
    
    This measure results in a value between $0$, completely unrelated profiles, and $1$, identical profiles.

    An alternative expression, mathematically equivalent but easier to understand for interpretability purposes, is found at~\cite{Brock1977}. We take this expression as our base Demographic Similarity ($DS$) measure:

    \begin{equation}\label{eq:ds}
        DS(X, X') = \text{R}(X, X') = 1 - 0.5\sum_{g\in G}\left|p_g - p'_g\right|\;.
    \end{equation}

    In the Jaccard family, the most notable examples are the quantitative Jaccard or Ruzicka similarity index~\cite{Ruzicka1958} and the Marczewski-Steinhaus similarity index~\cite{Marczewski1958}, sometimes called the Similarity Ratio. They can be generalized as:

    \begin{equation}
        \text{J}(X, X') = \frac{\sum_{g\in G} \min \left(n_g, n'_g\right)}{\sum_{g\in G} \max \left(n_g, n'_g\right)}\;.
    \end{equation}

    Both families obtain numerically different values when applied to normalized demographic profiles. However, the similarity values can be translated from one family to the other following $\text{J}(X, X')=\text{R}(X, X')/(2-\text{R}(X, X'))$. This relationship is non-linear but monotonic, and the resulting similarities generate the same ordering. Given this equivalence, we only consider one of the families. In particular, we chose the Renkonen-based $DS$ expression, which provides a more direct and easier to interpret explanation.

    In summary, DSAP will use $DS$ to measure the similarity between one of the demographic axis profiles derived from the input dataset and the reference profile. This will provide an easy-to-interpret value ranging between 0, complete dissimilarity, and 1, identical demographic profiles. In the next sections, we show how this methodology can be applied for different purposes, depending on the choice of reference profile.

\subsection{Demographic dataset comparison}\label{s:met_comparison}

    The first application of the DSAP methodology is the direct comparison of two real demographic profiles from two different datasets. As mentioned above, these profiles are compared on a given demographic axis, whose demographic axis profiles are compared. To make this comparison, the demographic profiles of both datasets must first be extracted following the first step of DSAP. Then, their similarity is computed with the proposed demographic similarity measure (Eq.~(\ref{eq:ds})).

    This type of pairwise comparison may be useful to obtain an overview of the datasets in a given problem, to analyze their demographic characteristics, and to understand their similarities and differences. For example, it can be used to support the combination of datasets to create mixed large-scale datasets~\cite{Gao2020}, to guide the selection of datasets to be used in sequential pretraining and fine-tuning methodologies~\cite{Raffel2020}, or to perform a general analysis and review of existing datasets in a field. 
    
    In this regard, we will study a large collection of FER datasets to show the usefulness of DSAP for this purpose. To do so, we propose using the demographic similarities computed by the methodology for each pair of datasets with a clustering method. In this way, we can better understand the properties shared by the demographic profiles of the FER datasets. More specifically, in our study we make use of a complete linkage clustering, thresholded over the cophenetic distance~\cite{Everitt2011}. The resulting clusters will correspond to groups of demographically similar datasets. The same methodology is extensible to any other field where DSAP can be applied.

\subsection{Bias measurement}\label{s:met_bias}

    The application of DSAP for bias measurement is based on the idea of comparing the demographic profile of a real dataset with ideal unbiased demographic profiles. Unlike other bias measures, DSAP will provide values close to 1 for balanced datasets (similar to the ideal one) and close to 0 for biased datasets. One of the advantages of DSAP is that we can use the same metric to measure different types of bias, obtaining the same range of output values with equivalent meanings, and therefore having better interpretability. Based on the DSAP methodology, we propose the following variants to measure different types of bias.

    \vspace{0.3em}\noindent\textbf{Representational bias ($DS_R$)}. Intuitively, the ideal reference for representational bias is the profile of a balanced dataset $X^\text{rep}$ with the same representation for all demographic groups on the corresponding axis. Knowing in advance the specific taxonomy that our auxiliary model produces, we can manually define the demographic profile for this ideal dataset by assigning a uniform representation of $1/\left|G\right|$ to each group in the demographic axis profile, with $\left|G\right|$ being the number of groups:

    \begin{equation}
        p^\text{rep}_g = \frac{1}{|G|}\;.
    \end{equation}
    
    Then, $DS_R$ for a population $X$ can be computed as:
    
    \begin{equation}
        DS_R(X) = DS(X, X^{\text{rep}})\;.
    \end{equation}

    Interestingly, we can manually adjust $X^\text{rep}$ to any desired target distribution, as long as $\sum_{g\in G}{p^\text{rep}_g}=1$ and $0 \leq p^\text{rep}_g \leq 1$ for all $g \in G$. This adjustment enables the relaxation of the representational bias definition, which can allow setting ideal profiles different from completely balanced ones, e.g., real-world representation profiles could be used, which was not possible with previous representational bias metrics.
    
    \vspace{0.3em}\noindent\textbf{Evenness ($DS_E$)}. Evenness is similar to representational bias, except for the fact that it does not consider groups that are not present in the dataset. Consequently, we generate a target demographic profile of an even dataset $X^\text{even}$ by assigning a $1/\left|G'\right|$ representation to each group represented in the dataset, with $\left|G'\right|$ being the number of groups represented, and 0 to those unrepresented:

    \begin{equation}
        p^\text{even}_g = \begin{cases}
            \frac{1}{|\left\{g\in G|p_g>0\right\}|} & \text{if }p_g > 0\\
            0 & \text{otherwise}\;.\\
        \end{cases}
    \end{equation}

    From this ideal profile, we then define $DS_E$ as:
    
    \begin{equation}
        DS_E(X) = DS(X, X^\text{even})\;.
    \end{equation}

    Again, we can manually adjust $X^\text{even}$ to any desired target profile. In this case, we require $\sum_{g\in G}{p^\text{even}_g}=1$, and additionally that $p^\text{even}_g = 0$ for all $g\in G$ such that $p_g=0$, and that $p^\text{even}_g > 0$ for the rest.
    
    \vspace{0.3em}\noindent\textbf{Stereotypical bias ($DS_S$)}. Finally, we can also measure the stereotypical bias between the demographic variables and the output variable. For this type of bias, we need to consider target classes. To do this, we propose using DSAP to measure the similarity between the demographic profile of the subset of the dataset labeled with a certain target class and the profile of the rest of the dataset. If the value $DS$ indicates a demographic difference for a given axis between the two profiles, that signals the presence of stereotypical bias. To obtain the bias value for the entire dataset, we average the similarity values for all classes. In an ideal unbiased dataset, all independent similarities should be 1, without any detectable difference between the profile of an individual class and the rest. In an extremely biased dataset with a completely different profile for each class, the similarities and the final result will be 0. Thus, $DS_S$ can be defined as:

    \begin{equation}
        DS_S(X) = \frac{\sum_{y\in Y}DS(X_y, X_{\hat{y}})}{|Y|}\;,
    \end{equation}
    where $Y$ is the set of target labels, and $X_y$ is the subset of the dataset labeled $y$, and $X_{\hat{y}}$ the rest of the dataset.

\subsection{Demographic dataset shift measurement}\label{s:met_shift}

    Finally, we can employ DSAP to measure demographic dataset shift. Recall that dataset shift corresponds to variations between the data used to train a model and either the data used to test it or the data found during deployment. 
    
    DSAP can be applied directly to measure demographic dataset shift between training and test partitions by comparing them demographically. That is, we can treat train and test partitions as independent datasets, calculating their individual demographic profiles, and computing the similarity between them on the different demographic axes. The similarity obtained can then be interpreted as the amount of shift between partitions. Similarity values close to 1 mean that both partitions share almost the same demographic profiles, whereas lower values would indicate the presence of varying amounts of demographic dataset shift, achieving the maximum amount at similarity values close to 0.

    Regarding the application of DSAP for detecting demographic dataset shift in deployment scenarios, additional considerations must be taken. In deployment, the model will receive a data stream, constituting an ever-growing dataset. If DSAP is periodically used to compare the train data against an increasingly large dataset, the impact of new data will be progressively reduced, eventually stabilizing the metric and making the detection of dataset shift difficult. To overcome this issue, we suggest fixing the maximum dataset size following a rolling window approach, a common strategy when dealing with data streams. In this way, the resulting rolling similarity measure will be comparable over time, becoming an appropriate tool for detecting demographic dataset shifts.

\section{Experimental framework}\label{s:casestudy}

    In this section, we present the implementation details of our study on FER using DSAP, whose results will be shown in Section~\ref{s:results}. First, Section~\ref{s:cs_fairface} presents the auxiliary demographic model that we use for the first step of DSAP, and the demographic classification associated to this model. Then, Section~\ref{s:cs_selection} presents the included datasets and their selection criteria. Finally, Section~\ref{s:cs_homogeneization} details the data preprocessing pipeline used to process the FER datasets before applying DSAP. The code associated with these experiments is available at \url{https://github.com/irisdominguez/DSAP}.

\subsection{Auxiliary demographic model} \label{s:cs_fairface}

    As the FER task is based on facial images, to obtain a demographic profile, as described in Section~\ref{s:met_profiling}, an appropriate auxiliary model is required. A good match for this task is FairFace~\cite{Karkkainen2021}, a model and dataset pair designed for the identification of age, gender and race from cropped face images. In the case of FER, FairFace matches the type of input for most datasets (close-up portrait images) and covers the most relevant potential demographic attributes. Both the model and the dataset are publicly available\footnote{\url{https://github.com/joojs/fairface}}. The original dataset used to train the model is made up of $108,501$ images gathered from the Flickr image hosting service and later manually labeled by external annotators. The model was compared with similar models trained on other demographic datasets, namely UTKFace, LFWA+, and CelebA, and showed good performance for the three demographic axes considered. Therefore, these will also be the demographic axes considered for our study. The labeling for the three axes is provided in the following labels:

    \begin{itemize}
        \item \textbf{Race} is classified into 7 possible groups, namely White, Black, Indian, East Asian, Southeast Asian, Middle Eastern, and Latino.
        \item \textbf{Gender} uses a simple binary categorization into Male or Female.
        \item \textbf{Age} axis is divided into 9 groups, namely $0$--$2$ years, $3$--$9$ years, $10$--$19$ years, $20$--$29$ years, $30$--$39$ years, $40$--$49$ years, $50$--$59$ years, $60$--$69$ years and $70+$ years.
    \end{itemize}

   Although the FairFace model and dataset are designed with bias issues in mind, i.e., by trying to gather a demographically diverse dataset, they are still limited. In particular, all samples are classified into one and only one group for each of the three axes, where the selection of groups limits the expressivity of the system and leaves out many possible identities on the axes of race and gender~\cite{Keyes2018}. Moreover, the labeling process for the original FairFace dataset relies on Amazon Mechanical Turk to annotate the demographic attributes of the subjects, which can introduce additional biases. Despite these shortcomings, FairFace is still an accurate and proven model for the task, and covers the main demographic attributes of interest for testing DSAP. As we use the same model for all datasets, the same potential biases apply to all datasets, improving the reliability of the final similarity score.
    
\subsection{Dataset selection} \label{s:cs_selection}

    For the case study, we consider the datasets presented in Table~\ref{table:datasets}. This list has been elaborated from a combination of dataset lists provided in reviews~\cite{Li2020}, datasets cited in various works, and datasets discovered by direct searches on the Internet. From all potential datasets, we select those that meet the following criteria: 

    \begin{enumerate}
        \item Datasets based on images or video datasets that can be used per frame. 3D, IR, and UV images are excluded from our analysis.
        \item Real image datasets. Although some artificial face datasets are available~\cite{Oliver2020}, we decide to exclude them, as demographic profiling can be unreliable in these cases.
        \item Datasets that include labels for the six basic emotions (anger, disgust, fear, happiness, sadness, and surprise) plus neutral, which is one of the most common encodings in FER datasets. For our purposes, the use of a single shared encoding makes stereotypical bias comparable between datasets, allowing the study of the $DS_S$ measure for stereotypical bias.
    \end{enumerate}

    \begin{table*}[ht]
      \begin{center}
        \caption{Summary of the FER Datasets and their characteristics.}
        \label{table:datasets}
        \resizebox{\textwidth}{!}{
        \pgfkeys{/pgf/number format/fixed}
        \begin{filecontents*}{data/datasets.csv}
Abbreviation,"Full Name",Year,Collection,Images,Videos,Subjects,Labelling$^a$
"ADFES \cite{vanderSchalk2011}","Amsterdam Dynamic Facial Expressions Set",2016,LAB,---,648,22,9
"AffectNet \cite{Mollahosseini2019}",AffectNet,2017,ITW,"291,652",---,---,"7 + N"
"CAER-S \cite{Lee2019}","CAER Static",2019,ITW-M,"70,000",---,---,"6 + N"
"CK \cite{Kanade2000}","Cohn-Kanade Dataset",2000,LAB,"8,795",486,97,"7 + N"
"CK+ \cite{Lucey2010}","Extended Cohn-Kanade Dataset",2010,LAB,"10,727",593,123,"7 + N + FACS"
"ExpW \cite{Zhang2017}","Expression in-the-Wild",2017,ITW,"91,793",---,---,"6 + N"
"FER2013 \cite{Goodfellow2013}","Facial Expression Recognition 2013",2013,ITW,"32,298",---,---,"6 + N"
"FERPlus \cite{Barsoum2016}","FER2013 Plus",2016,ITW,"32,298",---,---,"6 + N"
"GEMEP \cite{Banziger2012}","Geneva Multimodal Emotion Portrayals",2010,LAB,"2,817","1,260",10,17
"iSAFE \cite{Singh2020}","Indian Semi-Acted Facial Expression",2020,LAB,---,395,44,"6 + N + U"
"JAFFE \cite{Lyons1998, Lyons2021}","Japanese Female Facial Expression",1998,LAB,213,---,10,"6 + N"
"KDEF \cite{Lundquist1998}","Karolinska Directed Emotional Faces",1998,LAB,"4,900",---,70,"6 + N"
"LIRIS-CSE \cite{Khan2019}","LIRIS Children Spontaneous Facial Expression Video Database",2019,LAB,"26,000",208,12,"6 + U"
"MMAFEDB \cite{_MMAFACIALEXPRESSION}","Mahmoudi MA Facial Expression Database",2020,ITW,"128,000",---,---,"6 + N"
"MUG \cite{Aifanti2010}","Multimedia Understanding Group Facial Expression Database",2010,LAB,"70,654",---,52,"6 + N"
"NHFIER \cite{_NaturalHumanFace}","Natural Human Face Images for Emotion Recognition",2020,ITW,"5,558",---,---,"7 + N"
"Oulu-CASIA \cite{Zhao2011}",Oulu-CASIA,2008,LAB,"66,000",480,80,6
"RAF-DB \cite{Li2017, Li2019}","Real-world Affective Faces Database",2017,ITW,"29,672",---,---,"6 + N"
"SFEW \cite{Dhall2011}","Static Facial Expressions In The Wild",2011,ITW-M,"1,766",---,330,"6 + N"
"WSEFEP \cite{Olszanowski2015}","Warsaw Set of Emotional Facial Expression Pictures",2014,LAB,210,---,30,"6 + N + FACS"
\end{filecontents*}
\pgfplotstabletypeset[
            col sep=comma,
            text indicator=",
            columns={Abbreviation, 
                [index]1, 
                Year, 
                Collection, 
                Images, 
                Videos, 
                Subjects, 
                [index]7},
            sort=true,
            sort key=Year,
            string type,
            display columns/0/.style={column type = {l}},
            display columns/1/.style={column type = {l}},
            display columns/2/.style={column type = {r}},
            display columns/3/.style={column type = {l}},
            display columns/4/.style={column type = {r}},
            display columns/5/.style={column type = {r}},
            display columns/6/.style={column type = {r}},
            display columns/7/.style={column type = {l}},
            every head row/.style={
                before row=\toprule,after row=\midrule},
            every last row/.style={
                after row=\bottomrule}
        ]{data/datasets.csv}}
      \end{center}
     \footnotesize{$^a$ 6: angry, disgust, fear, sad, surprise, and happy. 7: 6 + contempt. N: Neutral. U: Uncertain. FACS: Facial Action Coding System.}
    \end{table*}


    The twenty datasets considered span from 1998 to 2020, with varied sizes (from $210$ to $291,652$ images) and different labeling and data characteristics. The data source in particular is a determinant factor when analyzing the demographic bias of these datasets~\cite{Dominguez-Catena2023a}, which can be classified into the following groups:

    \begin{itemize}
        \item \textbf{Laboratory-gathered} (\textit{Lab}) datasets, built under controlled conditions. They usually have high-quality and well-labeled images taken in a consistent environment, at the cost of a lower dataset size.
        \item ITW datasets from \textbf{Internet searches} (\textit{ITW-I}). These datasets use large amounts of images gathered from Internet image searches with manually established queries, resulting in highly variable quality. Usually, images are relabeled in a methodological fashion, which can improve the label accuracy.
        \item ITW datasets from \textbf{Motion Pictures} (\textit{ITW-M}). These datasets build on the success of ITW-I datasets, but employ images from motion pictures (TV shows and movies) to improve the image quality. These datasets tend to be a middle ground between Lab and ITW-I datasets in terms of dataset size and quality properties.
    \end{itemize}

\subsection{Dataset homogenization} \label{s:cs_homogeneization}

    To improve the quality of dataset comparison, it is important to ensure that profiling (Section~\ref{s:met_profiling}) is performed on images with characteristics as similar as possible. For our case study, this implies a series of homogenization steps to transform the original images into images compatible with our auxiliary model. The following steps are applied:

    \subsubsection{Frame extraction}
    
    Some of the datasets, namely ADFES, CK+, GEMEP, and iSAFE, are based on video. Fortunately, most of the videos in these Lab datasets tend to be of similar lengths and correspond to a single emotion, so for the purpose of demographic comparison, we can treat them as a single sample each. An exception to this pattern is found in the CK/CK+ datasets, where the emotion apex is represented only at the end of the sequence. As a result, we exclusively utilize the last 6 frames of each video in these two datasets.
    
    \subsubsection{Face extraction}

    For all datasets, we ensure that the image size and margin around the faces are homogeneous and as close as possible to those of the FairFace dataset used to create the auxiliary model. To this end, we replicate their setup~\cite{Karkkainen2021}, using the same Max-Margin (MMOD) CNN face extractor~\cite{King2015} implemented in DLIB\footnote{\url{http://dlib.net/}}. Face extraction is performed with a target face size of $224 \times 224$ and a margin around the face of $0.25$ ($56$ pixels), resizing the images as necessary. A zero padding (black border) is used when the resized image includes portions outside the original image.

    In the case of the EXPW dataset, which includes more than one face per image, we first use their provided face bounding boxes to isolate the faces and run the same face extractor on top of these images.

    \subsubsection{Subject aggregation}

    To facilitate the stability of the demographic profile generation, we use as much information as possible from each subject. In particular, in many Lab datasets, we can link multiple images to the same subject. For these datasets, we use all the available images of each subject, applying the FairFace model to each one of them, and aggregating the predicted values in each demographic axis. The final assigned demographic group is obtained by simple majority voting. Note that for ITW-I and ITW-M datasets this process cannot be performed as there is no identity information available.

\subsubsection{Label homogeneization}

    To facilitate the analysis of stereotypical bias, we ensure that all datasets have a consistent target labeling and the same number of labels. For datasets that have more than the required labels, we employ the subset of images corresponding to the six basic emotions~\cite{Ekman1971}: anger, disgust, fear, happiness, sadness, and surprise, plus a seventh category for neutrality.

\section{Results}\label{s:results}

    In this section, we test the usefulness of DSAP in the three proposed applications: demographic dataset comparison, bias measurement, and demographic dataset shift detection. To this end, we focus on a selection of 20 datasets for the FER problem. Consequently, we answer the following research questions:

    \begin{enumerate}
      \item Is DSAP useful for measuring demographic similarity between datasets?
      \item Does DSAP, in combination with clustering, help to understand the different types of bias in the datasets?
      \item Can DSAP be used to measure demographic bias? How does it compare with existing bias measures?
      \item Can DSAP be used to identify demographic data shift?
    \end{enumerate}

    The rest of this section is structured according to these questions, with each subsection answering one of them.

\subsection{Demographic similarity between datasets}\label{s:res_profiling}

\begin{figure}
        \centering
        \includegraphics[width=\columnwidth]{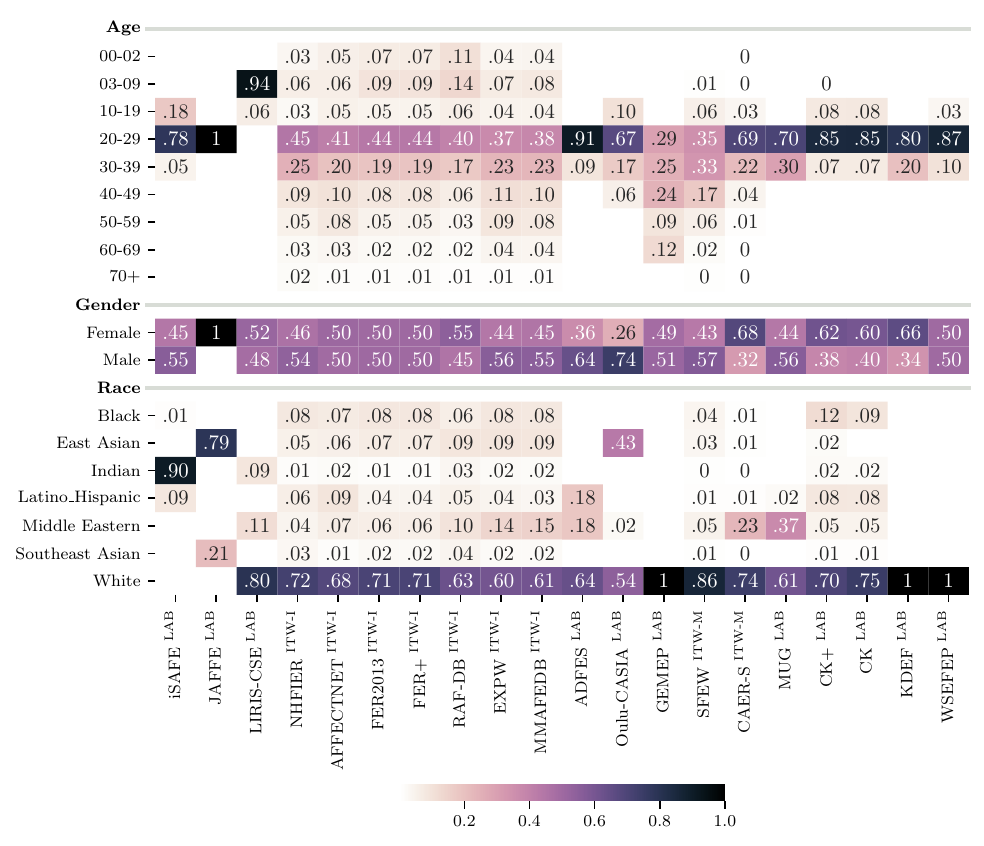}
        \caption{Demographic axis profiles of each of the datasets, as calculated from the FairFace model predictions. For each of the demographic axes, namely age, gender, and race, the proportion of subjects in each group is shown. In datasets where the identity of the subject is known, the demographic attributes are predicted for each sample and the mode of the sample predictions are considered for each subject.}
        \label{figure:demographic_distribution}
    \end{figure}

    As explained in Section~\ref{s:met_profiling}, we can employ an auxiliary model such as FairFace to obtain a comparable demographic profile between different datasets. The results of this process in the twenty FER datasets presented in Section~\ref{s:cs_selection} are summarized in Figure~\ref{figure:demographic_distribution}. Each column corresponds to one dataset and each horizontal section shows the profile on each demographic axis, namely, age, gender, and race. The combination demographic axis, created from all the possible combinations of groups in the other three axes, has a total of 126 unique groups and is not shown for the sake of space but is later considered in the DSAP application.

    It is important to note that the results of the demographic profiling are limited by the auxiliary model used. In this case, FairFace provides us with nine age segments (each covering a variable number of years), a binary gender classification, and seven racial groups. In the profiles extracted with FairFace, we can observe some inconsistencies with the information available for the datasets, such as the alleged presence of non-Indian subjects in iSAFE, which are excluded by design, or the presence of Southeast Asian subjects in JAFFE, a dataset composed only of Japanese women. These errors show some of the potential inaccuracies and biases of the demographic model itself, which motivates the use of the same model for all datasets so that the profiles have similar biases. Despite these issues, predictions match the distribution of the datasets with reasonable accuracy where demographic properties are known, such as the age profile of LIRIS-CSE (reported mean age $7.3$ years), the gender and race profiles of JAFFE (Japanese women), and the gender profile of Oulu-CASIA (reported to have $73.8\%$ male subjects). Further analysis of these profiles is provided in Section~\ref{s:res_clustering}.

\begin{figure}
        \centering
        \includegraphics[width=\columnwidth]{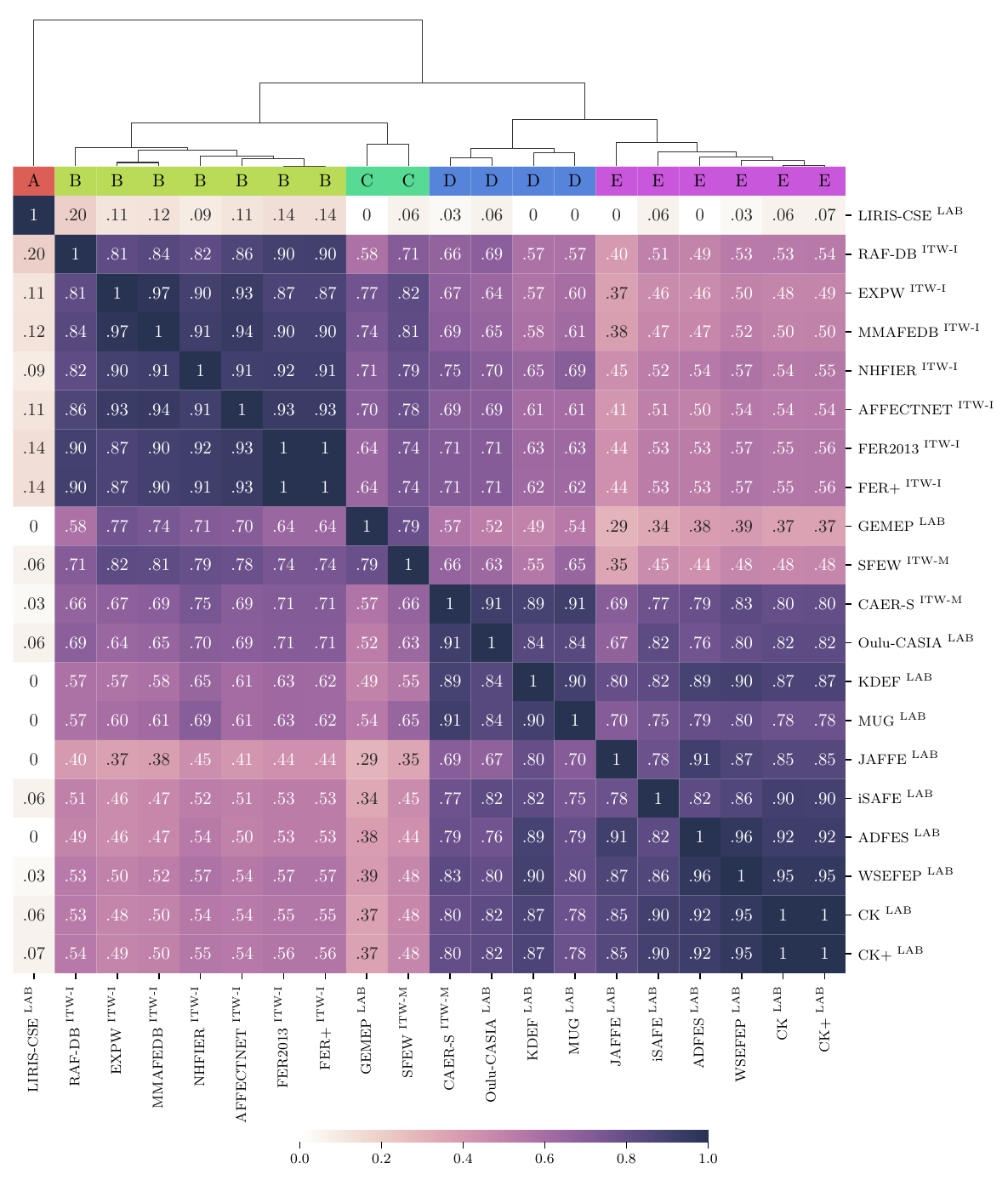}
        \caption{Demographic similarity comparison between the datasets in the age axis, and subsequent dataset clustering. The left-hand side dendogram is obtained from a complete linkage according to the similarity scores. The first column of the matrix shows a potential clusterization, obtained with a maximum cophenetic distance of $0.6$.}
        \label{figure:age_matrix}
    \end{figure}

\begin{figure}
        \centering
        \includegraphics[width=\columnwidth]{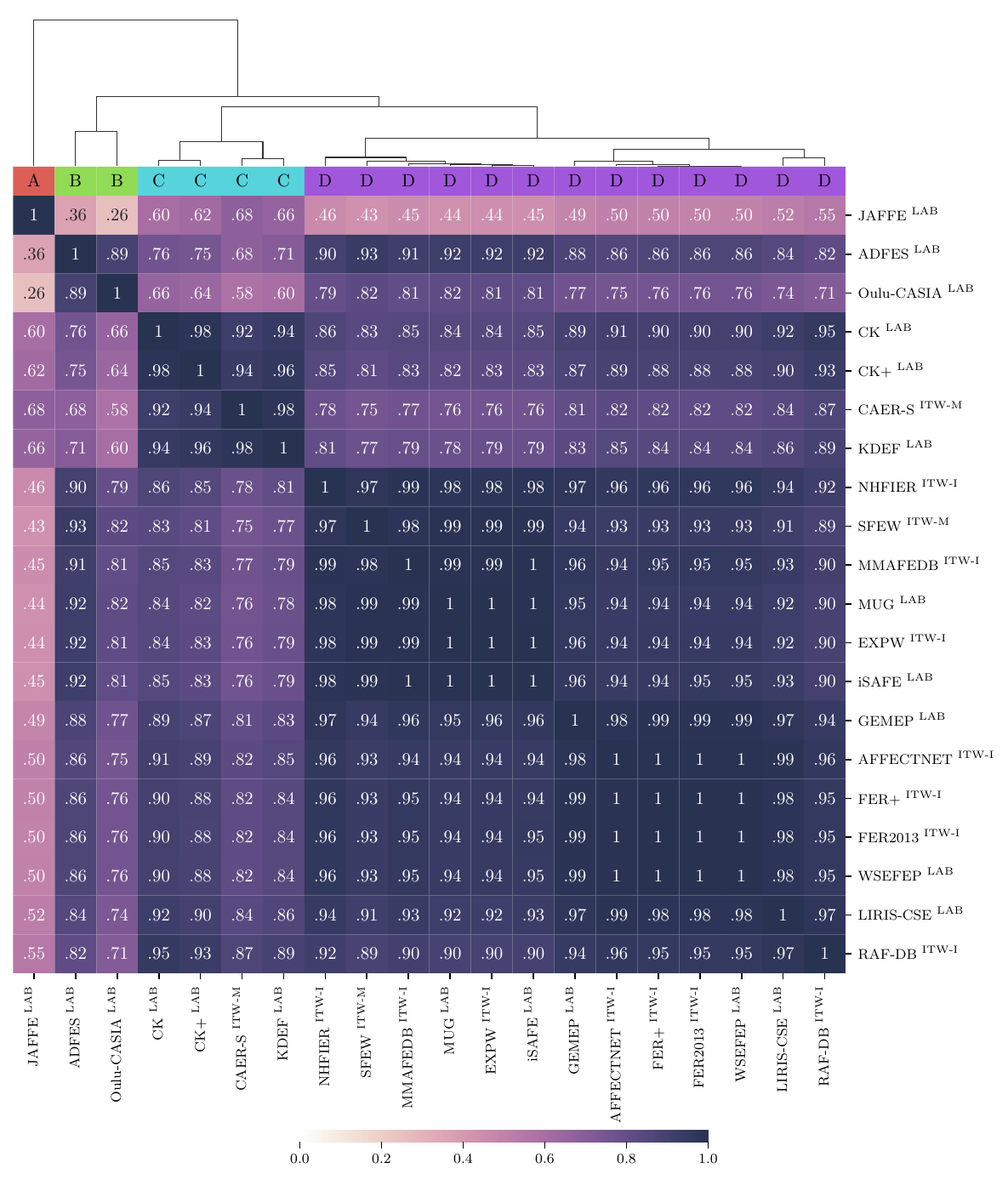}
        \caption{Demographic similarity comparison between the datasets in the gender axis, and subsequent dataset clustering. The left-hand side dendogram is obtained from a complete linkage according to the similarity scores. The first column of the matrix shows a potential clusterization, obtained with a maximum cophenetic distance of $0.6$.}
        \label{figure:gender_matrix}
    \end{figure}

\begin{figure}
        \centering
        \includegraphics[width=\columnwidth]{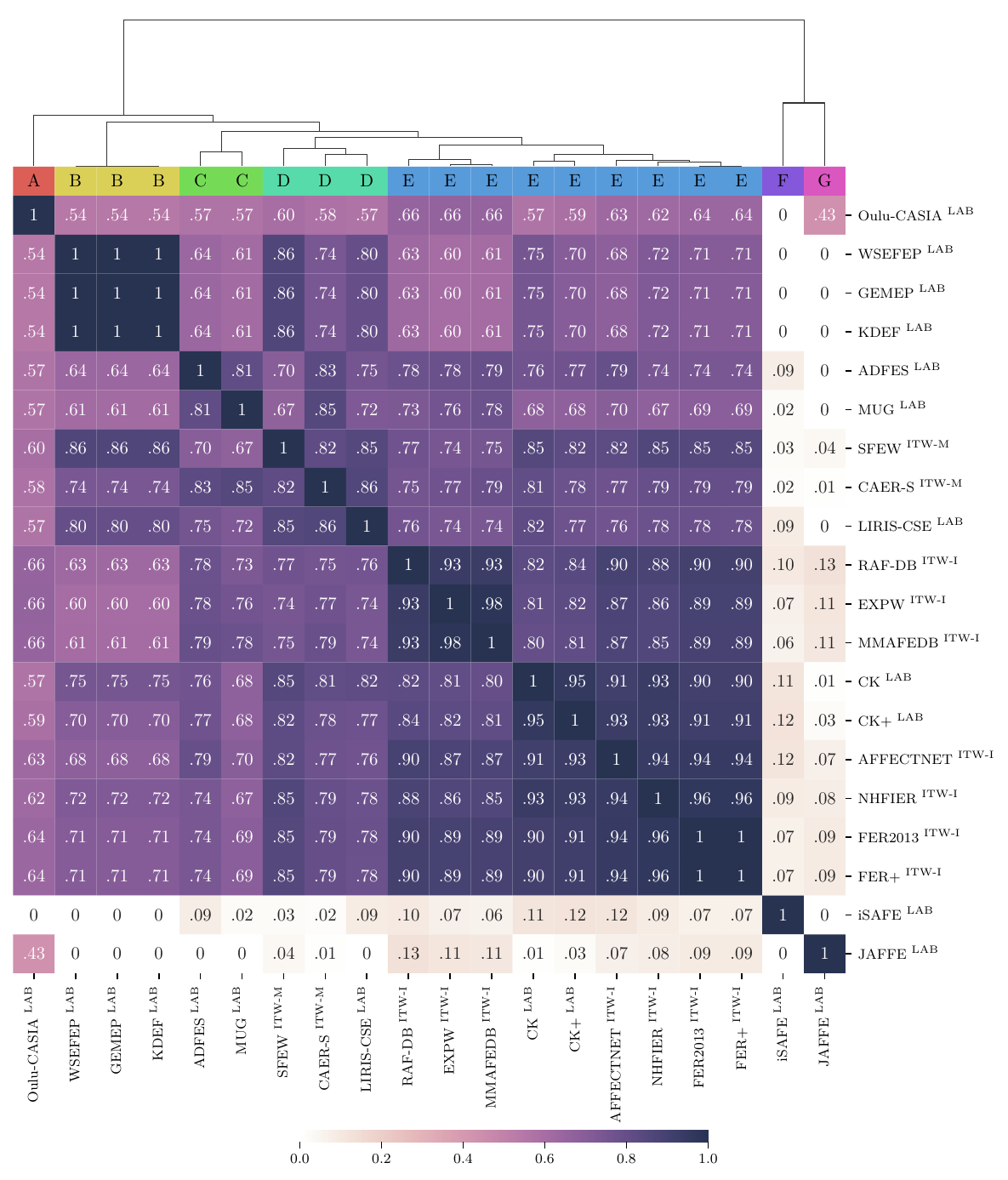}
        \caption{Demographic similarity comparison between the datasets in the race axis, and subsequent dataset clustering. The left-hand side dendogram is obtained from a complete linkage according to the similarity scores. The first column of the matrix shows a potential clusterization, obtained with a maximum cophenetic distance of $0.6$.}
        \label{figure:race_matrix}
    \end{figure}

\begin{figure}
        \centering
        \includegraphics[width=\columnwidth]{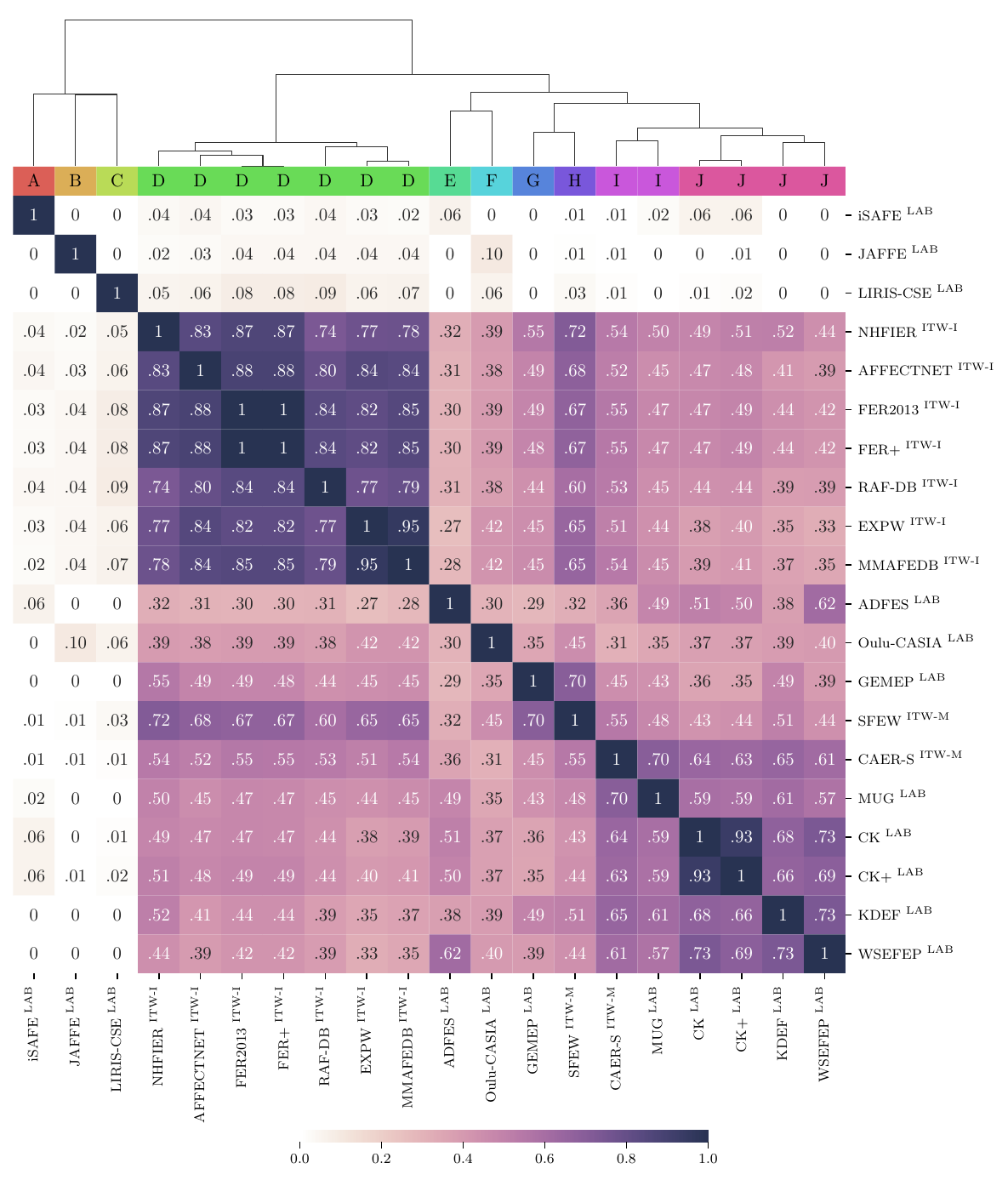}
        \caption{Demographic similarity comparison between the datasets in the combination demographic axis (age, gender and race), and subsequent dataset clustering. The left-hand side dendogram is obtained from a complete linkage according to the similarity scores. The first column of the matrix shows a potential clusterization, obtained with a maximum cophenetic distance of $0.6$.}
        \label{figure:combined_matrix}
    \end{figure}

    Figures~\ref{figure:age_matrix},~\ref{figure:gender_matrix},~\ref{figure:race_matrix} and~\ref{figure:combined_matrix} employ DSAP to compare the demographies of the selected datasets in pairs under a certain demographic axis (age, gender, race and combination, respectively). The upper portion of each of the figures corresponds to a cluster analysis based on these similarity scores, which will be discussed in detail in Section~\ref{s:res_clustering}.

    To compare the relationship between intuitive notions of similarity and the similarity values obtained by DSAP, we can focus on several dataset groups of interest. First, datasets collected under ITW conditions from Internet searches, such as NHFIER, AFFECTNET, FER2013, FER+, RAF-DB, EXPW, and MMAFEDB, all have intuitively similar demographic profiles on the three axes represented in Figure \ref{figure:demographic_distribution}. The differences between the proportions of the groups are under $0.08$ on the age axis, under $0.11$ on the gender axis, and under $0.12$ on the race axis. In the similarity matrices, the scores corresponding to these datasets are greater than $0.81$ on the age axis, greater than $0.94$ on the gender axis, and greater than $0.80$ on the race axis. On the combination axis (composed of 126 individual groups), these datasets obtain values greater than $0.74$, indicating high similarity.

    Other datasets, such as JAFFE and ADFES, can be very similar on a certain axis ($0.09$ maximum difference in age), while being very different on the other axis ($0.64$ and $0.79$ maximum differences in gender and race, respectively, with completely different race groups in both datasets). The similarity values in this situation vary accordingly, obtaining $0.91$ for age, $0.36$ for gender, and $0$ for race. In this case, the combination axis obtains a similarity value of $0$, since the individual subgroup composition of the two datasets is completely different, despite the similarities on the age axis. This behavior of the comparison on the combination axis is different from a mean of the similarities on the individual axes and can highlight key differences between datasets.

    \vspace{0.3em}\noindent\textbf{Conclusion}. Overall, we can observe that the demographic profiles obtained by the auxiliary model match the known demographic characteristics of the datasets, and that in the three individual demographic axes where an intuitive approach to demographic profile similarity is reasonable, the DSAP value follows accurately the intuitive notions of similarity. Therefore, we conclude that DSAP can be useful to measure demographic similarity between datasets.

\subsection{DSAP-based dataset clustering}\label{s:res_clustering}

    The similarity values reported in Figures~\ref{figure:age_matrix},~\ref{figure:gender_matrix},~\ref{figure:race_matrix}, and~\ref{figure:combined_matrix} can be used to establish clusters of demographically similar datasets. On the upper side of each of these figures there is a dendogram, produced from a complete linkage of the DSAP values. Additionally, the first row of each of the matrices shows a possible clustering assignment, obtained with a maximum cophenetic distance of $0.6$ (determined manually). In this section, we analyze each of the demographic axes individually.

    \vspace{0.3em}\noindent\textbf{Age}. In Figure~\ref{figure:age_matrix} we can observe the clustering in the age axis, resulting in 5 differentiated clusters. Cluster A identifies the most differentiated age dataset, the age-specific LIRIS-CSE, a dataset that consists only of children. Cluster B instead includes the FER datasets created from Internet images (ITW-I), which, as can be seen in Figure~\ref{figure:demographic_distribution}, all share a similar age profile, dominated by the age group $20-29$, but with a significant presence of the rest of the groups. GEMEP and SFEW form cluster C, with a strong and balanced representation of age groups $20-29$ and $30-39$. Finally, clusters D and E are characterized by a heavier dominance of the age group $20-29$, somewhat lower in cluster C ($67\%$ to $80\%$ of the subjects in the dataset) and higher in cluster D ($78\%$ to $100\%$ of the subjects in the dataset).

    \vspace{0.3em}\noindent\textbf{Gender}. The analysis in terms of gender is provided in Figure~\ref{figure:gender_matrix}. In this case, only four clusters are differentiated, as the datasets are more similar in this demographic axis. Specifically, cluster A includes only one dataset, JAFFE, composed only of women and thus not related to the others. Cluster B includes datasets dominated by female-presenting people ($64\%$ and $74\%$ of the dataset subjects), while cluster C instead includes datasets dominated by male-presenting people ($60\%$ to $68\%$ of the dataset subjects). Finally, cluster D collects most of the datasets that provide a balanced representation of both groups.

    \vspace{0.3em}\noindent\textbf{Race}. Figure~\ref{figure:race_matrix} shows the analysis in terms of race. In this case, up to 7 different clusters are observed, indicative of a larger demographic diversity between datasets. Three of these are trivial clusters composed of a single dataset (clusters A, F and G). From these isolated datasets, iSAFE (cluster F) and JAFFE (cluster G) are composed only of two of the commonly underrepresented racial groups, respectively, Indian and Japanese subjects, and show almost no relation to the rest of the datasets. The third one, Oulu-CASIA (cluster A), is composed of both apparently white and east Asian subjects, sharing some similarity with all datasets except iSAFE. Of the rest of the clusters, all relatively similar (similarity over $0.6$), cluster B collects datasets exclusively composed of white people, cluster C includes datasets with white and Middle Eastern representation, cluster D datasets with white subjects and some representation of the other groups, and finally cluster E collects datasets less dominated by the white race, which are mostly ITW-I datasets.

    \vspace{0.3em}\noindent\textbf{Combination}. Finally, Figure~\ref{figure:combined_matrix} compares the datasets based on the combination axis, where individual subgroups are considered for each combination of age, gender, and race. As expected, in this analysis, the clustering analysis highlights the differentiated datasets in any of the independent axes, generating up to 7 trivial clusters with a single dataset each (clusters A, B, C, E, F, G, and H). Of these clusters, it can be observed that three of them, namely A, B, and C, have very low similarities with any other dataset, while the other four, E, F, G, and H, still have some similarity to other datasets. This differentiates datasets that have completely distinct populations, not found in any other dataset, such as the A, B, and C clusters, from those that have only partially distinct populations, such as E, F, G, and H. Two smaller clusters, I and J, are also identified, characterized by mainly white populations in the age range $20-29$, mainly corresponding to laboratory-created datasets with a single ITW-M dataset as an exception. The last cluster, D, groups the ITW-I datasets, which are largely consistent with each other, even at the subgroup level, with very similar populations. 

    \vspace{0.3em}\noindent\textbf{Conclusion}. In general, this analysis highlights the broad similarity between the populations of the Internet search (ITW-I) datasets and shows how the other types of datasets, despite being individually more representationally biased and less diverse, can reach subgroups of populations that the ITW-I datasets do not. The DASP-based clustering analysis provides us with a general overview of the diversity or lack thereof in the datasets, highlighting which datasets cover unique demographic niches overlooked by others and which datasets are demographically redundant. This can guide the selection of datasets for training or testing purposes, as selecting at least one dataset from each cluster maximizes the inclusion of the available demographic groups.

\subsection{DSAP for bias measurement}\label{s:res_bias}

    Recall from Section~\ref{s:met_bias} that we can also employ DSAP as a measure of different types of bias in a dataset by comparing it with an ideal unbiased dataset. In this case, we will analyze the usefulness of DSAP derived measures to assess different types of biases and will compare them with existing bias measures.

\begin{figure*}
        \centering
        \includegraphics[width=.85\textwidth]{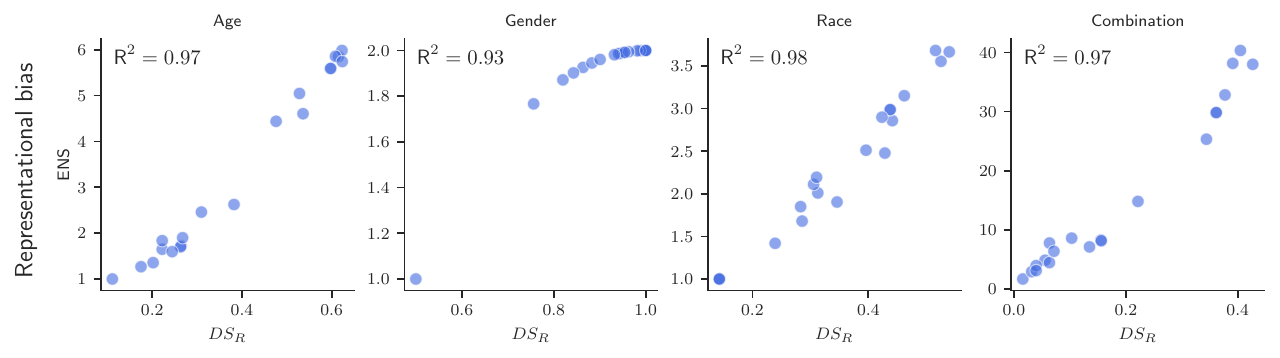}
        \includegraphics[width=.85\textwidth]{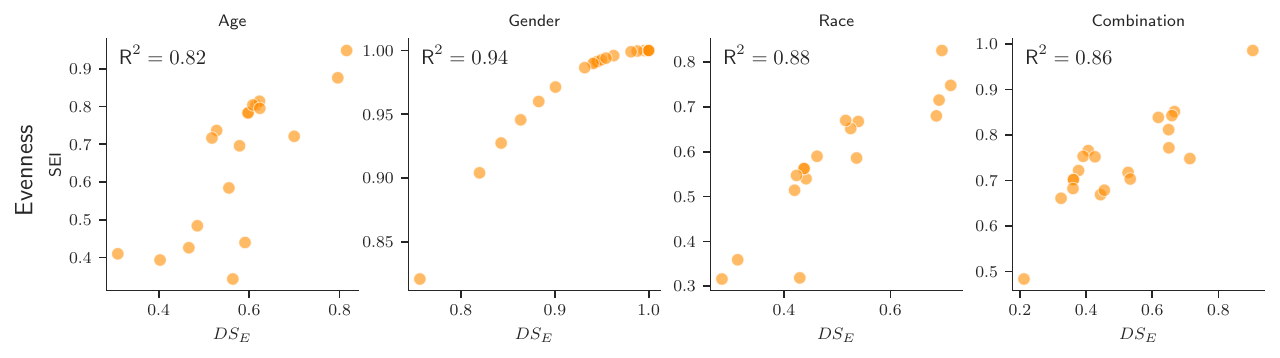}
        \includegraphics[width=.85\textwidth]{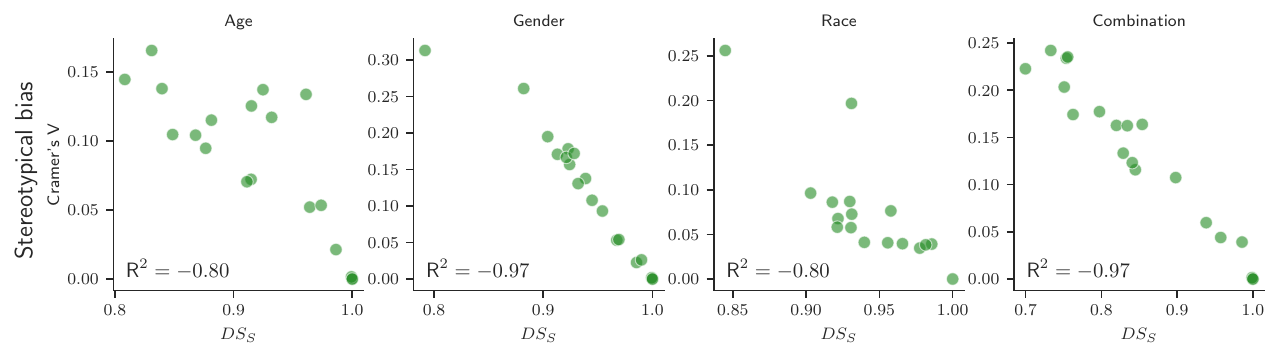}
        \caption{DSAP-based bias metrics (x-axis) compared to their classical counterparts (y-axis). In particular, the first row compares $DS_R$ and ENS, the second $DS_E$ and SEI, and the third one $DS_S$ and Cramer's V.}
        \label{figure:scatter_rep}
    \end{figure*}

    \vspace{0.3em}\noindent\textbf{Representational bias}. In the top row of Figure~\ref{figure:scatter_rep}, we can observe the relationship between $DS_R$ and the Effective Number of Species (ENS), a classical representational bias measure. The measures are calculated individually for each dataset and demographic axis. 

    Generally, a high correlation can be observed between the $DS_R$ and the ENS, with a $\text{R}^2$ value above $0.93$ in all cases, showing that the similarity measure closely matches a regular representational bias measure such as the ENS. The gender axis has only two groups, and thus a more clear definition of representational bias. In this case, the relationship on the gender axis seems nonlinear, while still being strictly monotonic. The non-linearity can be potentially related to the logarithmic formula of the ENS, but in any case does not affect the relative sorting of the datasets according to their bias.

    \vspace{0.3em}\noindent\textbf{Evenness}. As explained in Section~\ref{s:met_bias}, we can also use $DS_E$ as a measure of evenness. In the middle row of Figure~\ref{figure:scatter_rep}, we can observe the relationship between $DS_E$ and SEI, an evenness measure. In this case, the correlation is weaker, but still significant, as measured by $\text{R}^2$ above $0.82$. As with $DS_R$ and ENS, we can observe a less linear relationship in the gender axis, while still being strictly monotonic.

    \vspace{0.3em}\noindent\textbf{Stereotypical bias}. Finally, the bottom row in Figure~\ref{figure:scatter_rep} compares $DS_S$, following the method described in Section~\ref{s:met_bias}, with another measure of stereotypical bias, Cramer's V. It is important to note that, in this case, the relationship with V is inverse, since V measures bias directly, while $DS_S$ measures similarity to an ideal, being higher for unbiased datasets. The relationship between $DS_S$ and V is weaker than that of representational bias, but the absolute value of $\text{R}^2$, above $0.8$ in all axes, still shows a significant correlation. In this case, the relationship is stronger and more linear for the combination and gender axes, while the race and age axes have weaker relationships.

\begin{figure}
        \centering
        \includegraphics[width=0.9\columnwidth]{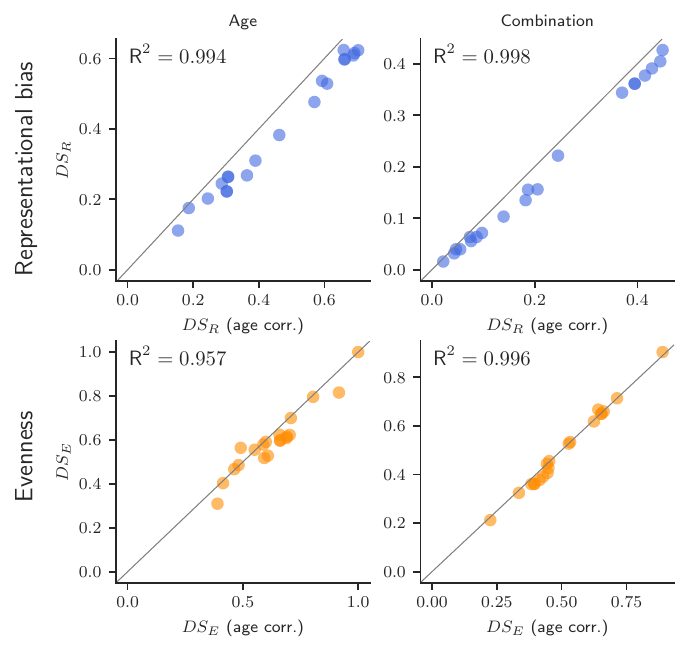}
        \caption{Comparison of the age-corrected $DS_E$ and $DS_R$ (x-axis) with their uncorrected variants (y-axis) in the age and combination axes (the rest of the axes are unaffected). The age-corrected target axis profile was created following the 2021 world age distribution.}
        \label{figure:scatter_corrected}
    \end{figure}

    \vspace{0.3em}\noindent\textbf{Age correction}. In Figure~\ref{figure:scatter_corrected} we compare the results of the DSAP-based representational and evenness metrics when using a corrected ideal target profile. As explained in Section~\ref{s:met_bias}, $DS_R$ and $DS_E$ allow one to correct the ideal demographic axis profiles for specific target populations. We show an example of this by adjusting the age axis profile to the age distribution of the world (2021 data~\cite{_WorldPopulationProspects}), which affects the results on the age and the combination axes for both of these metrics.
    
    In the top row of Figure~\ref{figure:scatter_corrected} we can observe that this correction increases the $DS_R$ value for all datasets in both the age and combination axes. This increased value indicates a lower bias, since the most underrepresented age groups in the datasets are also the least represented in the world. The bottom row of Figure~\ref{figure:scatter_corrected} shows the case of $DS_E$, where the impact of the correction is weaker, with most datasets obtaining similar values after the correction. This can be explained by the fact that, for the age axis, many datasets simply do not represent most of the age groups, especially the lower and upper age groups. As these groups are also the least represented in the world population, in the $DS_R$ computation they have a greater impact, while for the $DS_E$ they are ignored, and the final result does not show a significant change.

    \vspace{0.3em}\noindent\textbf{Conclusion}. These results support the usage of DSAP-based bias measures ($DS_R$, $DS_E$, and $DS_S$) as a viable replacement for previous dataset bias measures. Unlike previous measures, all DSAP-based bias measures share a range between 0 and 1, with a consistent meaning for these limits across datasets and types of biases. In particular, a value of 1 always represents total equivalence to a theoretical unbiased dataset, and a value of 0 represents total dissimilarity to the ideal dataset and, therefore, maximum bias.

\begin{figure*}
        \centering
        \includegraphics[width=0.95\textwidth]{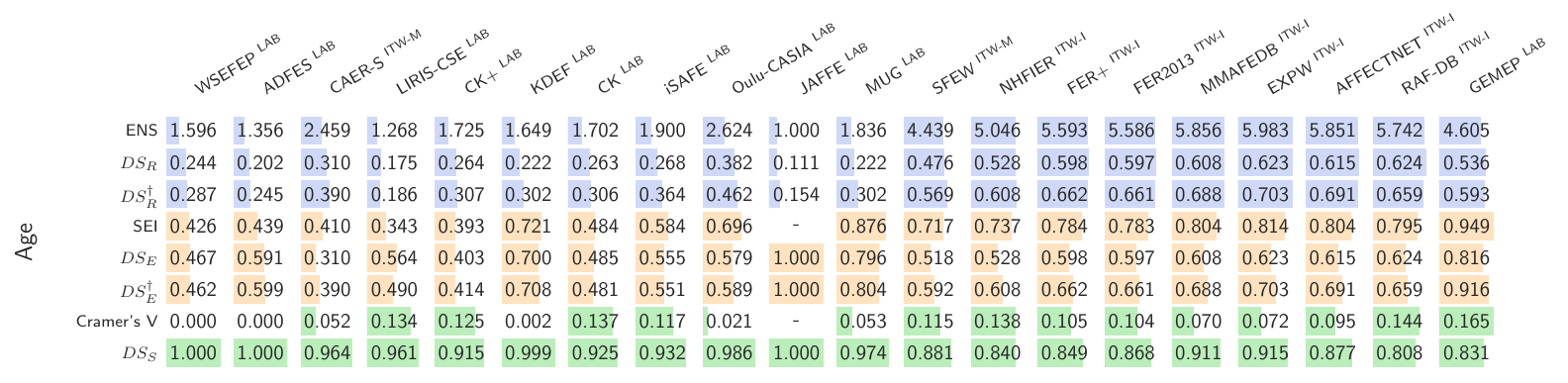}\label{figure:bias_alternative_a}
        
        \includegraphics[width=0.95\textwidth]{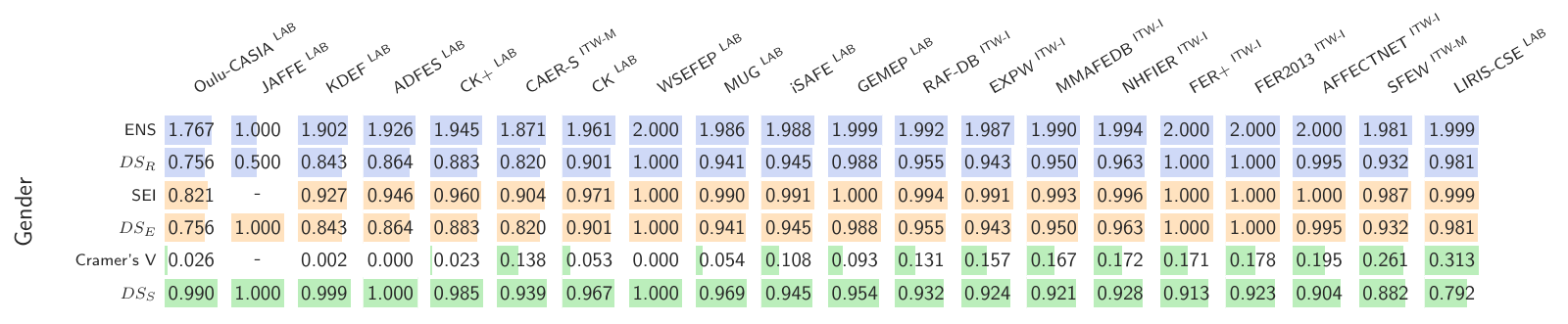}\label{figure:bias_alternative_b}
        
        \includegraphics[width=0.95\textwidth]{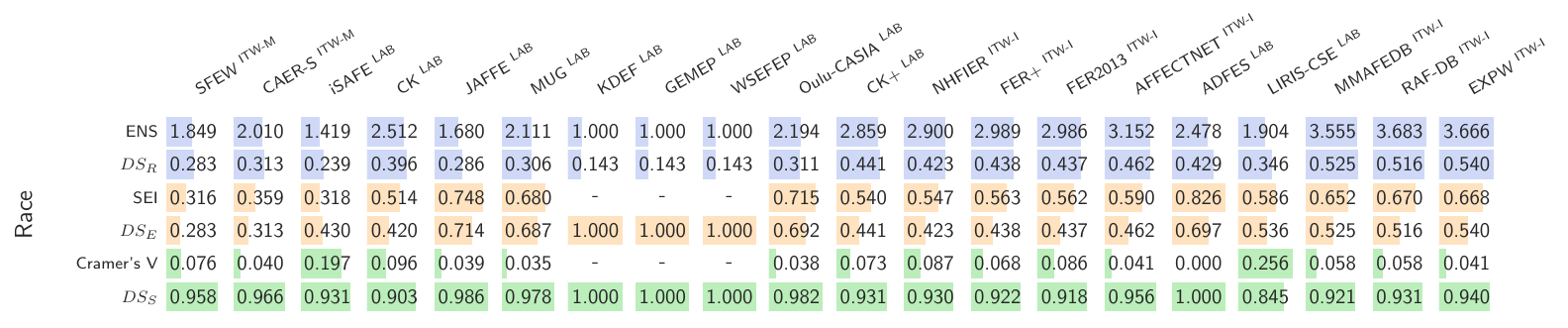}\label{figure:bias_alternative_c}
        
        \includegraphics[width=0.95\textwidth]{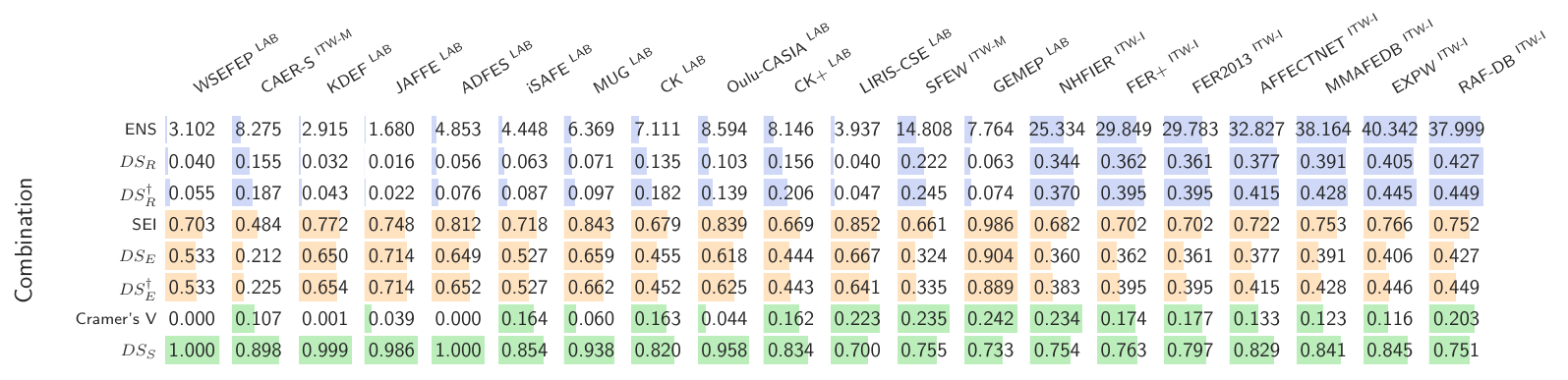}\label{figure:bias_alternative_d}
        \caption{Detailed results of each measure (rows) and dataset (columns) in the four axes. Measures (rows) indicated with $\dagger$ are calculated assuming the world age distribution as the ideal distribution.}
        \label{figure:bias_alternative}
    \end{figure*}

    \vspace{0.3em}\noindent\textbf{Detailed analysis}. Figure~\ref{figure:bias_alternative} include the detailed results, where we can observe the outliers and particular cases. In this figure, we employ different colors for the representational bias measures (ENS, $DS_R$ and $DS_R$ corrected by age), the evenness measures (SEI, $DS_E$ and $DS_E$ corrected by age), and the stereotypical bias measures (Cramer's V and $DS_S$).

    First, in the age, gender, and race results we can observe several cases where datasets with a single represented group leave both the SEI and Cramer's V value undefined. However, the corresponding $DS_E$ and $DS_S$ values are well defined and directly interpretable, with a value of $1$, the maximum similarity to the ideal datasets. This is in agreement with the intuition that single-group datasets cannot be uneven or stereotypically biased.

    In the evenness analysis, we can observe some discrepancies, especially in the age axis. A good example is LIRIS-CSE ($\text{SEI} = 0.343$, $DS_E = 0.564$) and iSAFE ($\text{SEI} = 0.584$, $DS_E = 0.555$), which share a close $DS_E$ score but a distinct SEI value. In Figure~\ref{figure:demographic_distribution}, we can observe the demographic profiles of these two datasets, where we can observe how iSAFE has three represented groups (with a representation of $78\%$, $18\%$ and $5\%$, respectively) and LIRIS-CSE only two (representations of $94\%$ and $6\%$). These situations are not directly comparable, and the general intuition of evenness does not indicate which should be considered more or less even. Thus, different mathematical definitions of evenness, such as SEI and $DS_E$, can be expected to yield different results.

    An additional interesting point of comparison between the DSAP-based bias measures and other bias measures is that, by using the same measure for different types of biases, the scores are more comparable among them. For example, previous work~\cite{Dominguez-Catena2023a} found that stereotypical bias was generally weaker than representational bias in FER datasets. Although this result is impossible to assess when comparing only Cramer's V and ENS, the DSAP-based measures seem to support it, with a very high $DS_S$ value in all axes (global average at $0.91\pm0.076$), which indicates a lower stereotypical bias, and a significantly lower $DS_R$ value (global mean $0.46\pm0.3$), which indicates a higher representational bias.

    Finally, another key advantage of DSAP-based bias measures is the possibility of using a corrected ideal. In these results, the comparison between using an ideal balanced demographic axis profile and an age-corrected one reveals a consistent impact on the representational bias and evenness scores. In particular, when comparing the balanced ideal with the age-corrected ideal in representational bias, we observe a $0.063 \pm 0.022$ increase in $DS_R$ values in the age axis and a $0.027 \pm 0.012$ increase in the combination axis. These increments indicate that when using the age-corrected ideal, the amount of bias is lower, as the datasets are closer to this relaxed realistic ideal. When performing the same comparison for evenness, we also observe an increase of $0.035 \pm 0.044$ in the $DS_E$ values of the age axis and an increase of $0.011 \pm 0.018$ in the combination axis, which further reinforces the general reduction in bias. As expected, in both the representational bias and the evenness analysis, the variations due to age correction are larger in the age axis than in the combination axis, where the age has a more limited influence.

    In general, we can observe a good correlation between DSAP-based bias measures and previous bias measures. The exceptions to this correlation, mostly in the cases of evenness and stereotypical bias, are generally explainable as variations in the mathematical definitions of these types of bias. $DS$-based bias measures seem to have several advantages over the classical measures, such as shared range and interpretation between different types of biases and the support of novel analysis based on corrected ideals.





\subsection{DSAP for demographic dataset shift}\label{s:res_shift}

\begin{figure}
        \centering
        \includegraphics[width=\columnwidth]{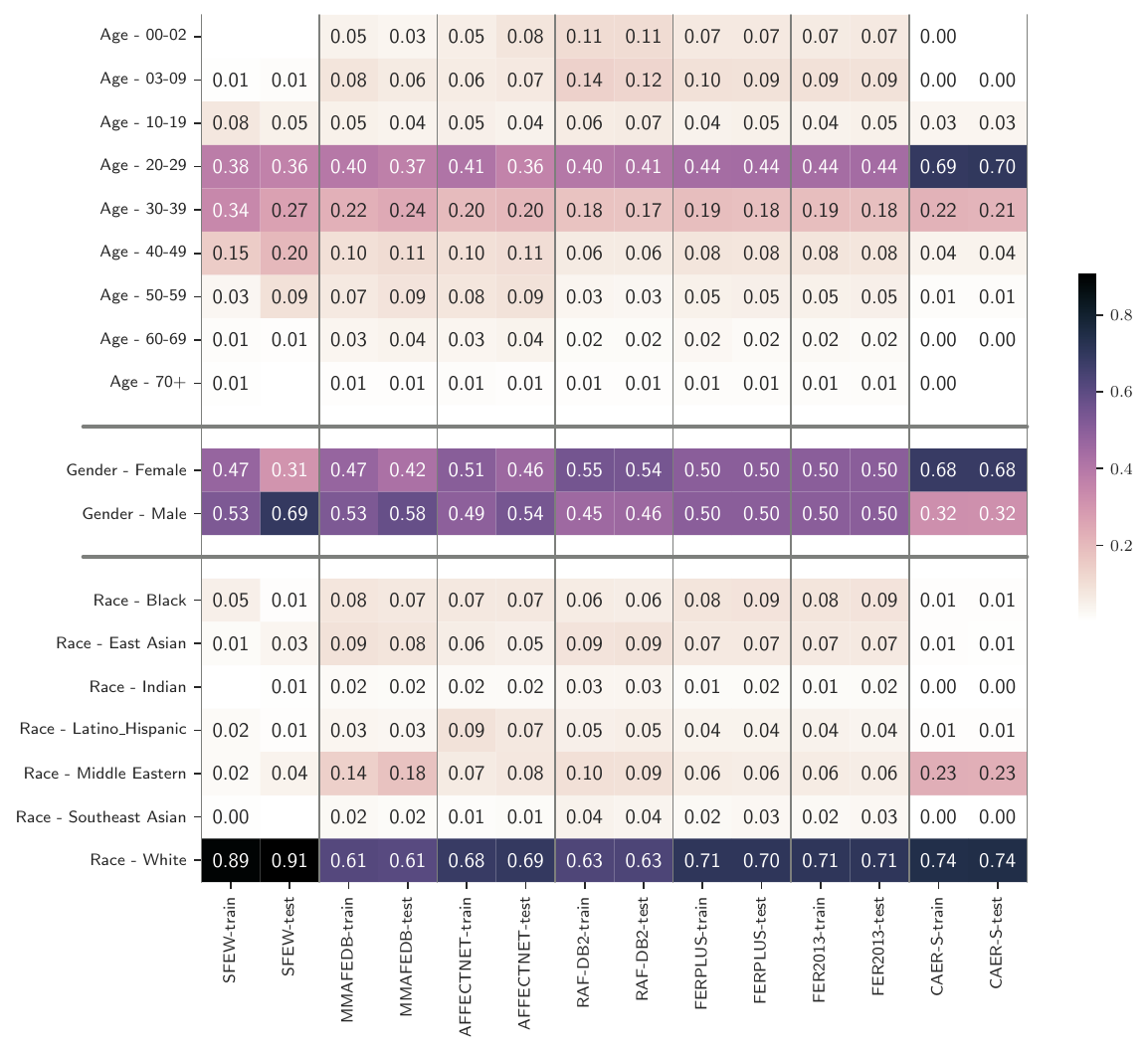}
        \caption{Demographic profiles of the train and test partitions of the seven datasets with predefined partitioning.}
        \label{figure:drift_demographic_distribution}
    \end{figure}

    To illustrate the application of DSAP for demographic dataset shift detection, in Figure~\ref{figure:drift_demographic_distribution} we focus on seven datasets that provide a predefined test or validation partition, which could be subject to demographic dataset shift. As we did for all datasets in Section~\ref{s:res_profiling}, we use FairFace to obtain a demographic profile for each partition (train and test independently) of the seven datasets.

    For most datasets, we can observe a similar profile in both partitions, maintaining the same representational biases. For example, CAER-S has a strong bias in the gender axis, with $68\%$ of the dataset being female-presenting people, but since both partitions maintain the same bias, this is not indicative of demographic dataset shift. On the contrary, SFEW seems to have a strong shift, especially in the age and gender axes. We can observe that the test partition is more representationally biased than the train partition on the gender axis, with $69\%$ of the included subjects being male, while only $53\%$ of them are male in the train partition. On the age axis, the demography also varies, although in this case it seems to be less biased on the test partition.

\begin{figure}
        \centering
        \includegraphics[width=0.7\columnwidth]{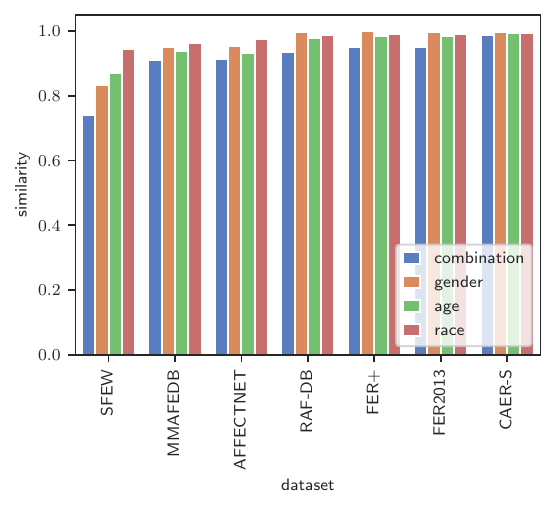}
        \caption{Results of the $DS$ measure applied as a demographic dataset shift measure, comparing the demographic profiles of the train and test partition of the datasets that provide them.}
        \label{figure:drift_measure}
    \end{figure}

    Figure~\ref{figure:drift_measure} shows the results of using DSAP to compare the training and testing profiles of each dataset. The same shifts can be observed in this case, summarized in a single, interpretable value for each axis. We can observe a relatively high shift in the SFEW dataset, characterized by low similarity, where the gender and combination axes, in particular, seem to vary significantly between the partitions. In contrast, the rest of the datasets have a high similarity value, greater than $0.9$ on all axes, indicating a low amount of data shift.

    In general, DSAP serves as a straightforward indicator of demographic dataset shift in fixed train and test partitions. Following the indications in Section~\ref{s:met_shift}, it can also be applied to detect demographic dataset shift in deployed systems. In both of these scenarios, DSAP provides a simple and reliable indicator that can be monitored in production scenarios where manually evaluating the demographic profile changes is unfeasible, especially for demographic axes with a high number of groups or problems with multiple possible demographic axes.

\section{Conclusion and Future Work}\label{s:conclusion}

    In this work, we have proposed DSAP, a two-step methodology for demographic comparison of datasets. This kind of comparison can be useful when creating new datasets and when working with several datasets simultaneously, as a way of detecting demographic blindspots, common grounds, and selection biases shared across a wide range of datasets and situations.

    Furthermore, DSAP can be applied to bias measurement by comparing real datasets with ideal unbiased versions. In this regard, we proposed three variants adapted for different types of demographic bias, namely representational bias ($DS_R$), evenness ($DS_E$), and stereotypical bias ($DS_S$). When comparing these DSAP variants with other bias measures, DSAP results in a homogeneous range with a straightforward interpretability, shared between the three types of bias. Additionally, unlike previous measures, DSAP-based measures can be adapted to specific target populations, relaxing bias definitions to more accurate and realistic standards.

    As a third application of DSAP, it can be applied to the problem of detecting demographic dataset shift, where the training dataset has different demographic characteristics from the data used for testing or the data found when deploying a system. In these situations, which can heavily impact the model performance and fairness, DSAP provides a simple measure that can be monitored for shift detection.

    To show the validity of DSAP, we applied it to a set of twenty datasets designed for the FER task. In these datasets, our methodology allowed us to identify groups of demographically similar datasets and single out those that cover demographic blindspots. Furthermore, we confirm previously found demographic biases, adding novel perspectives such as the comparison across types of biases and showing how the bias scores change when relaxing the ideals to more realistic requirements. Finally, we used DSAP to identify potential demographic dataset shift in the published datasets, finding significant variations between train and test partitions in a well-known dataset, SFEW.

    Some potential lines of future work are left open. Along with the application of DSAP to new datasets and contexts to further assess its capabilities, the current implementation has limitations that could be improved. In particular, for the demographic profiling step, new demographic models with more fine-grained classification of demographic groups, or allowing multiple group labels per subject, could improve the quality of the profiles obtained. Alternative auxiliary models targeting other demographic properties, such as socio-economic status, could unlock new perspectives on existing datasets. Furthermore, the consideration of ordinal or continuous variables could be explored as ways to extend DSAP to new applications. Supported by some of these extensions, new applications of DSAP should be explored, such as the identification of the best datasets or subsets for model training or directed dataset combination to create maximally diverse or unbiased datasets.

\FloatBarrier
\bibliographystyle{IEEEtran}
\bibliography{23_dataset_comparison,FER_bibtex}

\begin{IEEEbiography}[{\includegraphics[width=1in,height=1.25in,clip,keepaspectratio]{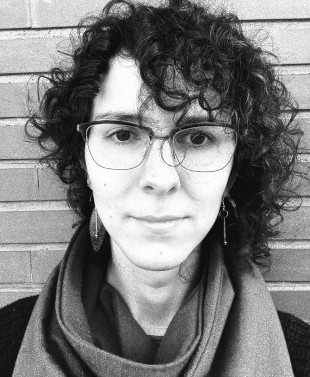}}]{Iris Dominguez-Catena} 
received the B.Sc. and M.Sc. degrees in computer science from the Public University of Navarra, in 2015 and 2020, respectively. She worked for several private software companies from 2015 to 2018. She is currently a Ph.D. candidate with the Public University of Navarra, working on demographic bias issues in Artificial Intelligence. Her research interests focus on AI fairness, bias detection and mitigation, and other ethical problems of AI deployment in society.
\end{IEEEbiography}

\begin{IEEEbiography}[{\includegraphics[width=1in,height=1.25in, clip, keepaspectratio]{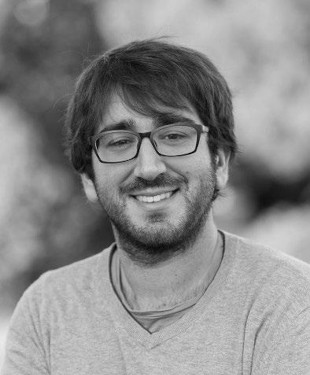}}]{Daniel Paternain}
received the M.Sc. and Ph.D. degrees in computer science from the Public University of Navarra, Pamplona, Spain, in 2008 and 2013, respectively. He is currently Associate Professor with the Department of Statistics, Computer Science and Mathematics. He is also the author or coauthor of almost 40 articles in journals from JCR and more than 50 international conference communications. His research interests include both theoretical and applied aspects of information fusion, computer vision, and machine learning.
\end{IEEEbiography}

\begin{IEEEbiography}[{\includegraphics[width=1in,height=1.25in,clip,keepaspectratio]{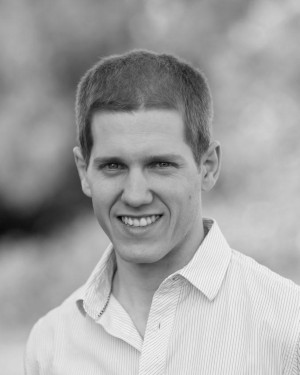}}]{Mikel Galar} (M14)
received the M.Sc. and Ph.D. degrees in computer science from the Public University of Navarra, Pamplona, Spain, in 2009 and 2012, respectively. He is currently an Associate Professor at the Public University of Navarra. He is the author of 50 published original articles in international journals and more than 80 contributions to conferences. He is a co-author of a book on imbalanced datasets and a book on large-scale data analytics. His research interests are  machine learning, deep learning, ensemble learning and big data. He  received the extraordinary prize for his PhD thesis from the Public University of Navarre and the 2013 IEEE Transactions on Fuzzy System Outstanding Paper Award (bestowed in 2016).


\end{IEEEbiography}

\vfill

\end{document}